\documentclass[sigconf]{acmart}
\AtBeginDocument{%
  \providecommand\BibTeX{{%
    \normalfont B\kern-0.5em{\scshape i\kern-0.25em b}\kern-0.8em\TeX}}}

\copyrightyear{2021}
\acmYear{2021}
\setcopyright{acmlicensed}\acmConference[KDD '21]{Proceedings of the 27th
ACM SIGKDD Conference on Knowledge Discovery and Data Mining}{August
14--18, 2021}{Virtual Event, Singapore}
\acmBooktitle{Proceedings of the 27th ACM SIGKDD Conference on Knowledge
Discovery and Data Mining (KDD '21), August 14--18, 2021, Virtual Event,
Singapore}
\acmPrice{15.00}
\acmDOI{10.1145/3447548.3467453}
\acmISBN{978-1-4503-8332-5/21/08}

\usepackage{enumitem}       % for no indent of itemize. with leftmargin=*
\usepackage{graphicx}       % required for the \raisebox command
\usepackage{bm}             % bold the number
\usepackage{wrapfig}        % make figure float
\usepackage{makecell}       % makecell
\usepackage{xcolor}         % for the \note{} command

\def\imagetop#1{\vtop{\null\hbox{#1}}}

\acmSubmissionID{rst1725}

\begin{document}
\fancyhead{}

\title{How Interpretable and Trustworthy are GAMs?}


% \author{Chun-Hao Chang$^{1,2,3}$, Sarah Tan$^{4}$, Ben Lengerich$^{5,6}$, Anna Goldenberg$^{1,2,3}$, Rich Caruana$^{7}$}
% \affiliation{%
%   \institution{$^1$University of Toronto, $^2$Vector Institute, $^3$Hospital of Sick Children\\ $^4$Cornell University, $^5$MIT, $^6$Broad Institute, $^7$Microsoft Research}
% }
% \email{kingsley@cs.toronto.edu, ht395@cornell.edu, blengeri@mit.edu}
% \email{anna.goldenberg@utoronto.ca, rcaruana@microsoft.com}

\author{Chun-Hao Chang}
\affiliation{%
  \institution{University of Toronto, Vector Institute, Hospital of Sick Children}
 \country{}
}
\email{kingsley@cs.toronto.edu}

\author{Sarah Tan}
\affiliation{%
  \institution{Cornell University}
 \country{}
  }
\email{ht395@cornell.edu}

\author{Ben Lengerich}
\affiliation{%
  \institution{MIT, Broad Institute}
 \country{}
}
\email{blengeri@mit.edu}

\author{Anna Goldenberg}
\affiliation{%
  \institution{University of Toronto, Vector Institute, Hospital of Sick Children}
 \country{}
}
\email{anna.goldenberg@utoronto.ca}

\author{Rich Caruana}
\affiliation{%
 \institution{Microsoft Research}
 \country{}
 }
\email{rcaruana@microsoft.com}

\renewcommand{\shortauthors}{Chang et al.}

\begin{abstract}
% Generalized additive models (GAMs) have become a leading model class for discovering bias and pattern in data.
Generalized additive models (GAMs) have become a leading model class for interpretable machine learning.
However, there are many algorithms for training GAMs, and these can learn different or even contradictory models, while being equally accurate.
Which GAM should we trust?
In this paper, we quantitatively and qualitatively investigate a variety of GAM algorithms on real and simulated datasets. 
We find that GAMs with high feature sparsity (only using a few variables to make predictions) can miss patterns in the data and be unfair to rare subpopulations.
Our results suggest that inductive bias plays a crucial role in what interpretable models learn and that tree-based GAMs represent the best balance of sparsity, fidelity and accuracy and thus appear to be the most trustworthy GAM models.
\end{abstract}





\maketitle
\section{Introduction}

As the impact of machine learning on our daily lives continues to grow, we have begun to require that ML systems used for high-stakes decisions (e.g., in healthcare, finance and criminal justice) not only be accurate but also satisfy other properties such as fairness or interpretability~\citep{doshi2017towards, lipton2018mythos}.
Generalized additive models (GAMs) have emerged as a leading model class that is designed to be accurate, and yet simple enough for humans to understand and mentally simulate how a GAM model works~\citep{doctor2020interpretable}, and is widely used in scientific data exploration~\citep{hastie1995generalized, pedersen2019hierarchical, Izadi2020generalized} and model bias discovery~\citep{tan2018distill, tan2018learning}.

\setlength\tabcolsep{0.pt} % default value: 6pt
\begin{figure}[tbp]
  \begin{center}

\begin{tabular}{ccc}
   & (a) Race
   & (b) Length of Stay
   \\
    \raisebox{-3.5\normalbaselineskip}[0pt][0pt]{\rotatebox[origin=c]{90}{\small Log odds}}
   & \imagetop{\includegraphics[width=0.52\linewidth,height=0.40\linewidth]{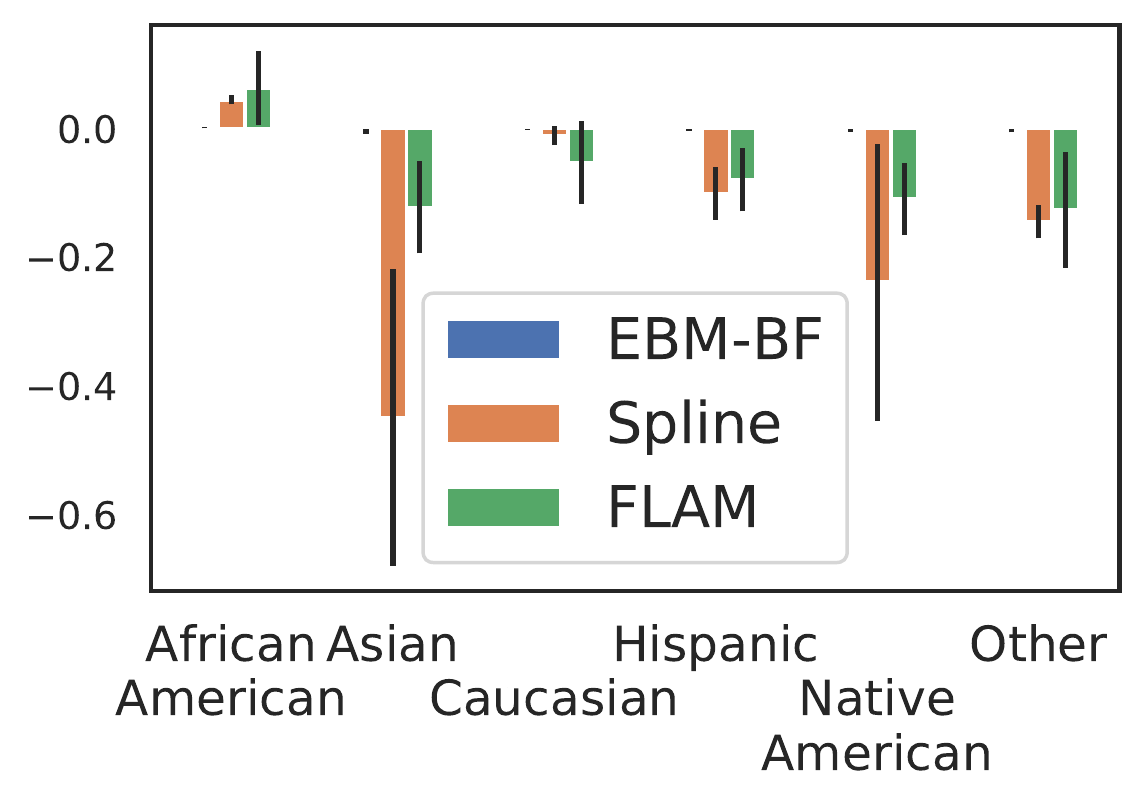}}
   & \imagetop{\includegraphics[width=0.48\linewidth,height=0.35\linewidth]{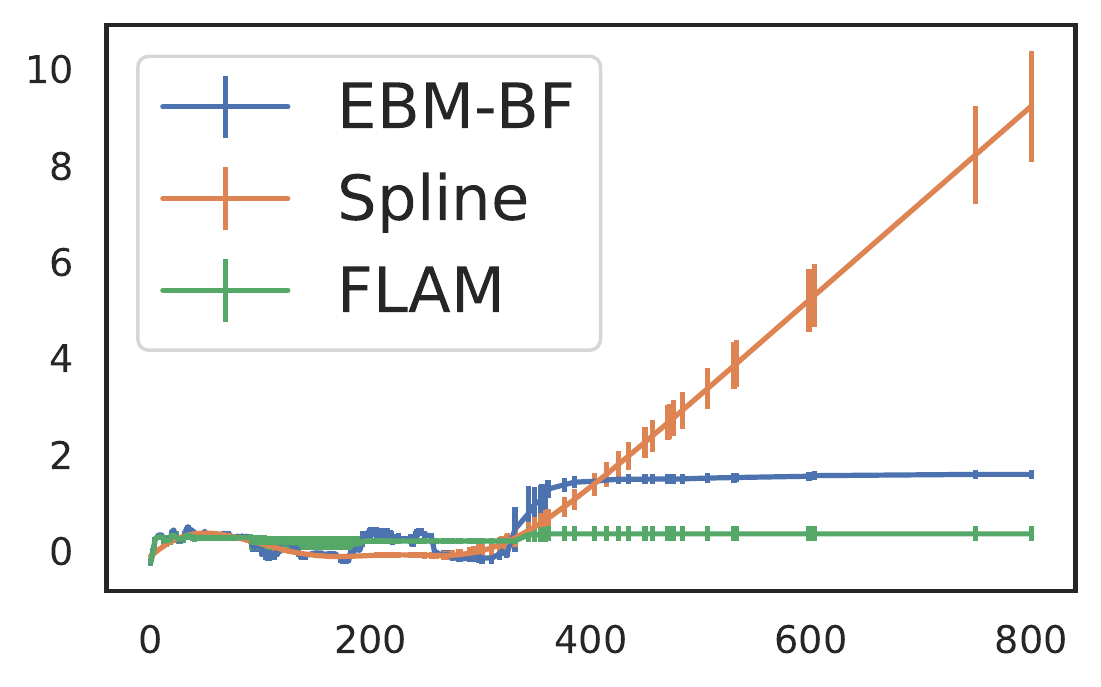}}
 \\
  \end{tabular} 
\end{center}

  \caption{
     Three GAMs with similar accuracy trained on the COMPAS recidivism dataset (two of six features shown) that tell very different stories about the dataset. 
     On the left, EBM-BF disagrees with FLAM and Spline about the presence of racial bias in the data. On the right FLAM suggests that length of stay  has no impact on risk, but Spline shows that risk grows strongly with length of stay.}
  \label{fig:poster_child}
\end{figure}

GAMs were originally trained using smoothing splines \citep{hastie1990generalized, wahba1990spline} that enforced smoothness in the learned functions. Later, several trend-filtering based methods including fused lasso additive models were proposed to make learned functions more sparse and jumpy \citep{fusedlasso, sadhanala2017additive}. \citet{lou2012intelligible} also proposed using boosted-tree-based methods to fit GAMs. Subsequent work showed the value of tree-based GAMs on two healthcare datasets \citep{caruana2015intelligible}, and also to help audit black-box models to ensure fairness~\citep{tan2018distill}.

\begin{figure*}[tbp]
  \begin{center}

\begin{tabular}{lccccc}

   & (a) EBM, EBM-BF
   & (b) XGB, XGB-L2
   & (c) Spline
   & (d) FLAM
   & (e) Strawmen \\ \toprule
    
  & Age & Age & Age & Age & Age \\
 \raisebox{3.\normalbaselineskip}[0pt][0pt]{\rotatebox[origin=c]{90}{\small Log odds}} 
 & \includegraphics[width=0.2\linewidth]{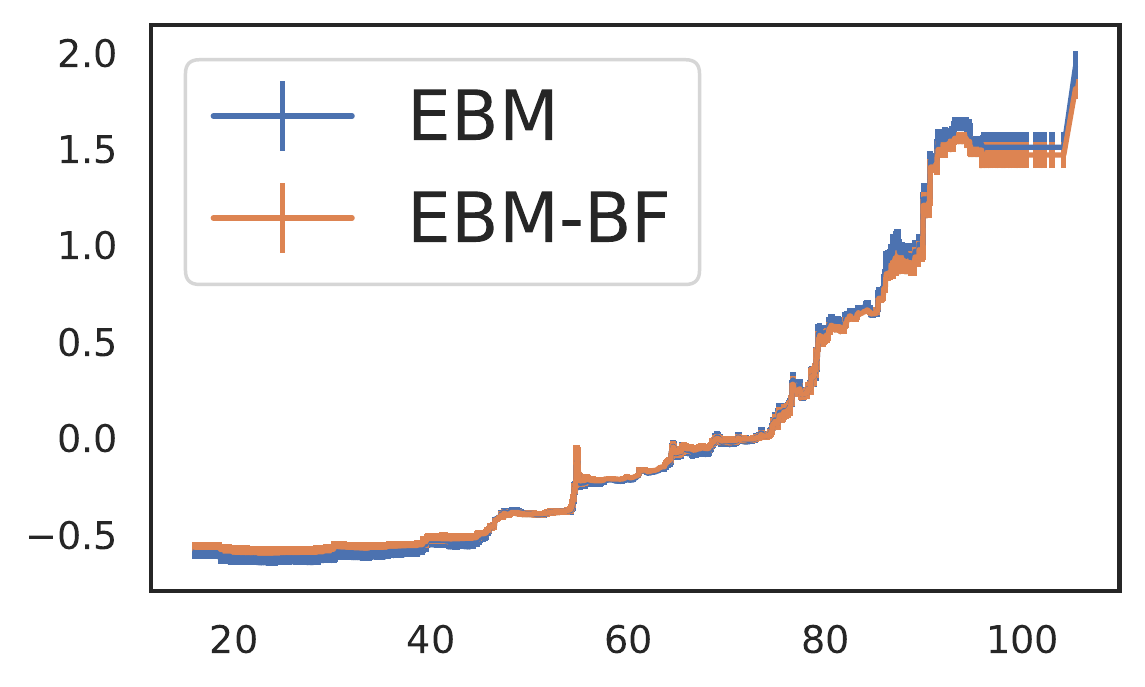}
 & \includegraphics[width=0.2\linewidth]{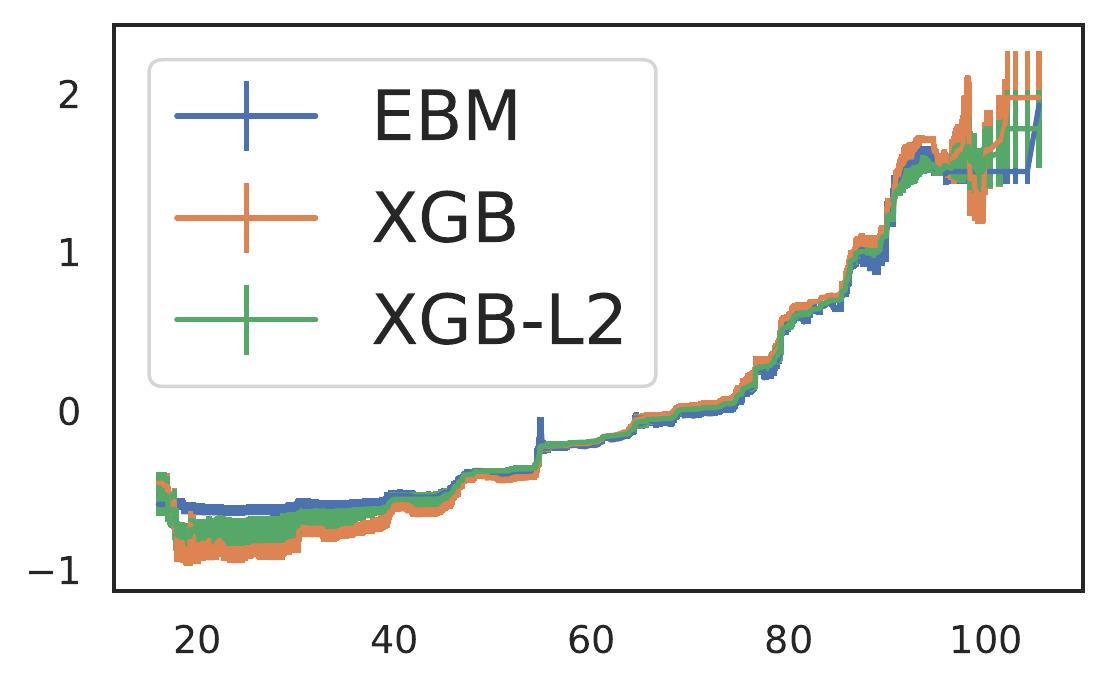}
 & \includegraphics[width=0.2\linewidth]{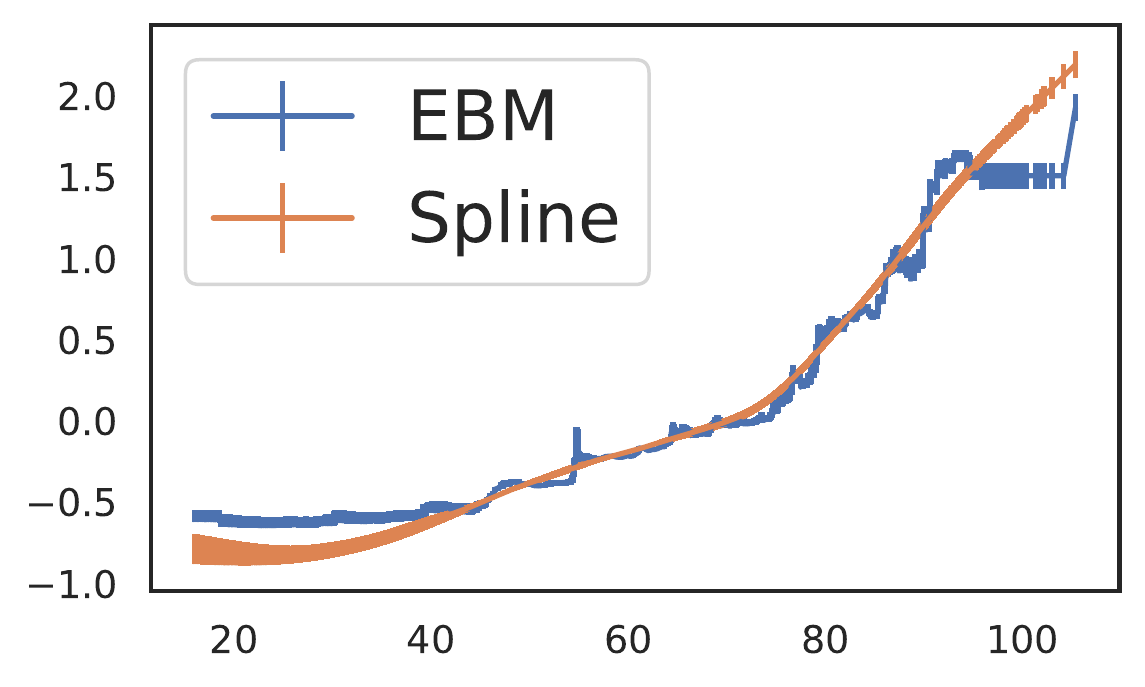}
 & \includegraphics[width=0.2\linewidth]{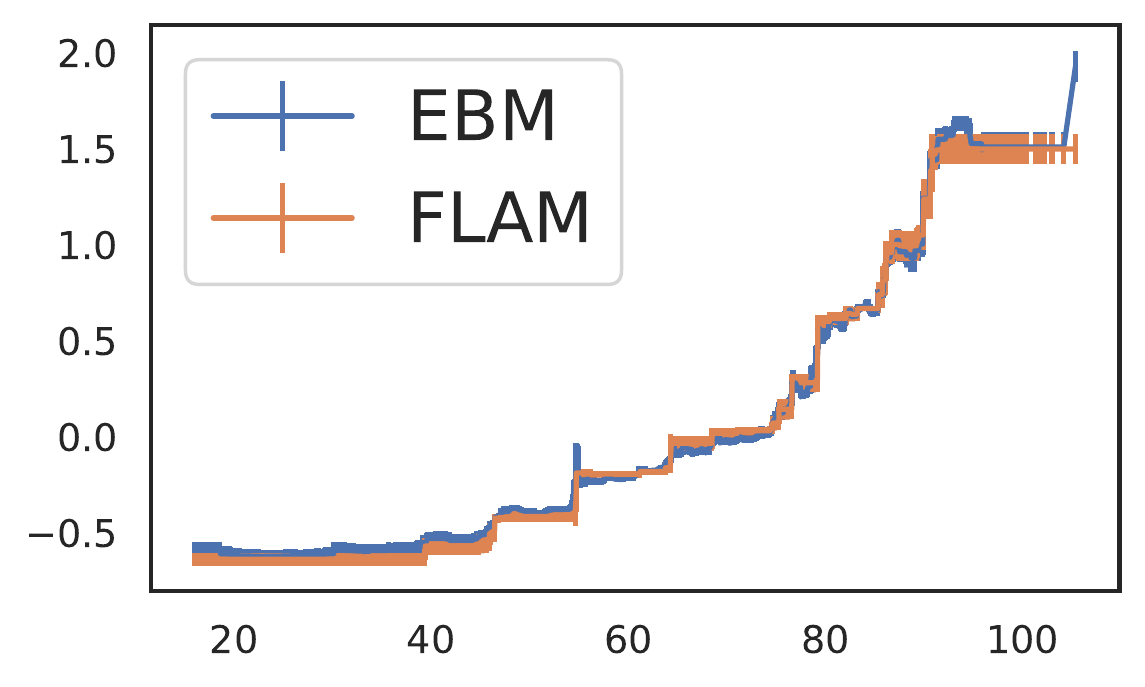}
 & \includegraphics[width=0.2\linewidth]{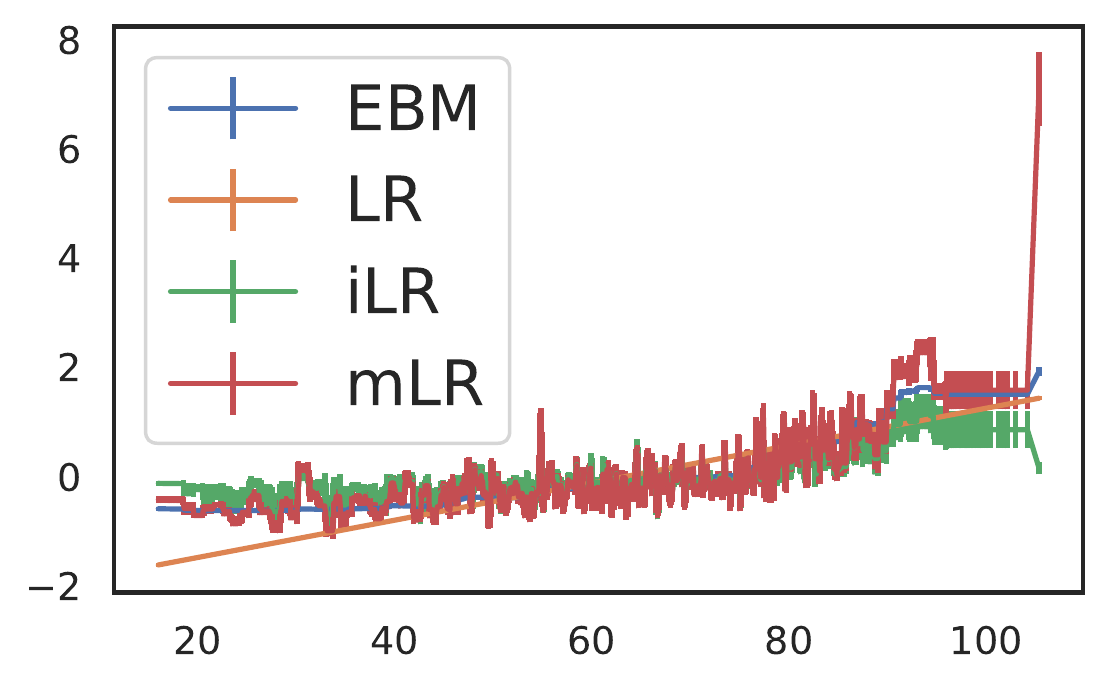} 
 \\
 
  & Systolic BP & Systolic BP & Systolic BP & Systolic BP & Systolic BP \\
  \raisebox{3.\normalbaselineskip}[0pt][0pt]{\rotatebox[origin=c]{90}{\small Log odds}} 
 & \includegraphics[width=0.2\linewidth]{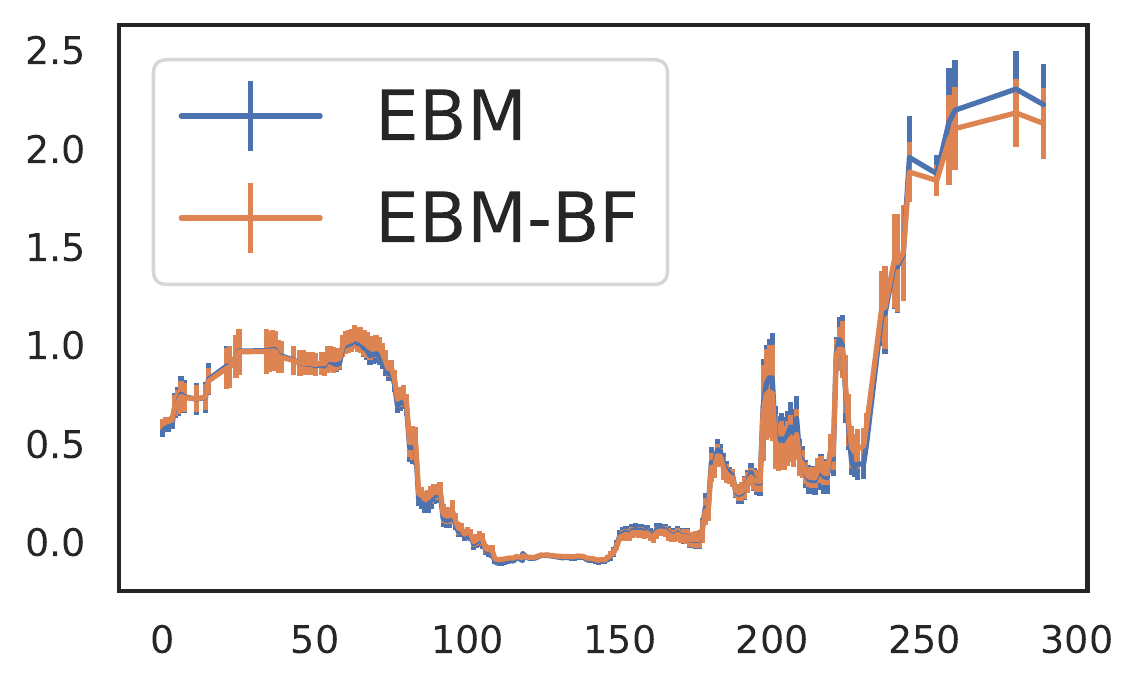}
 & \includegraphics[width=0.2\linewidth]{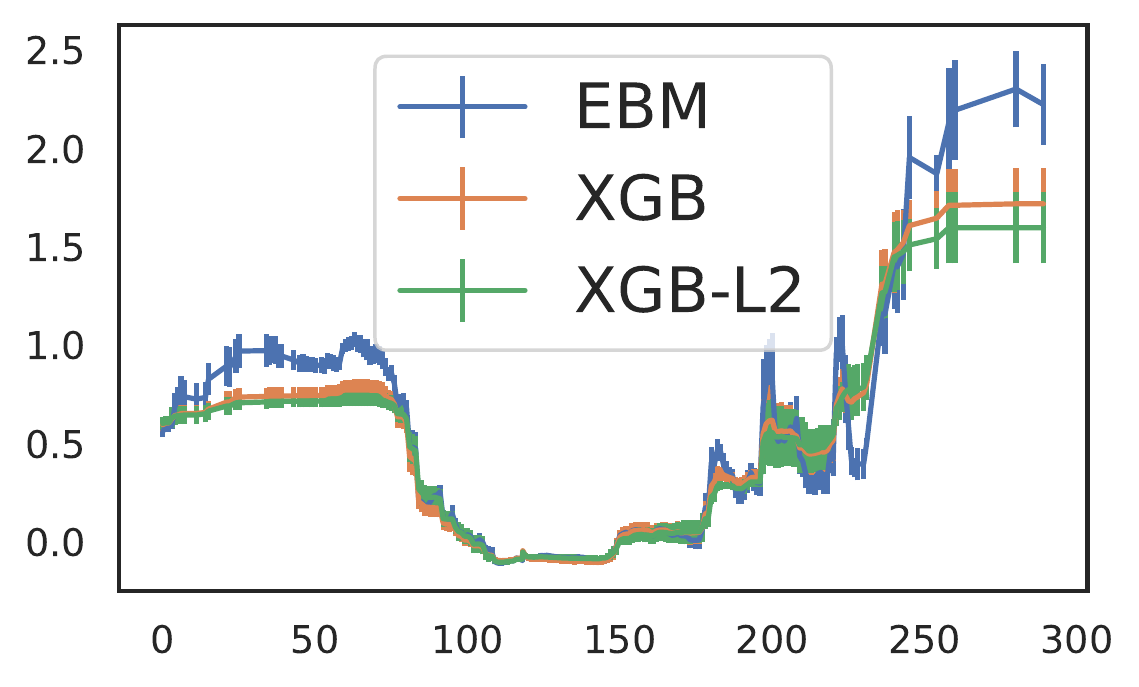}
 & \includegraphics[width=0.2\linewidth]{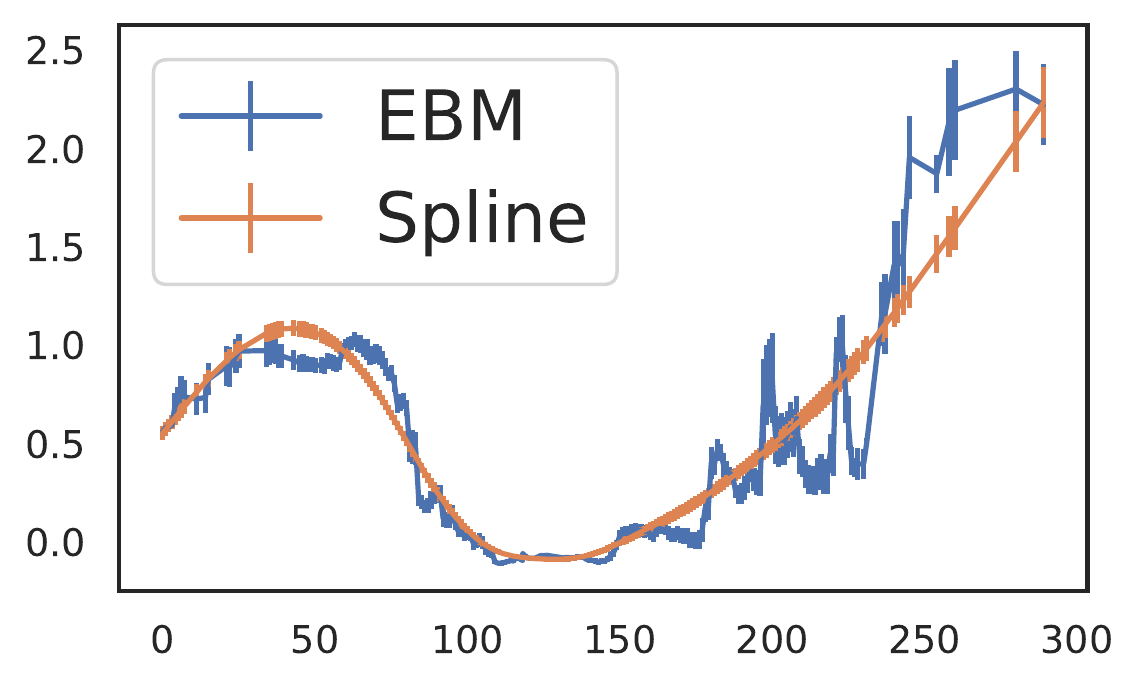}
 & \includegraphics[width=0.2\linewidth]{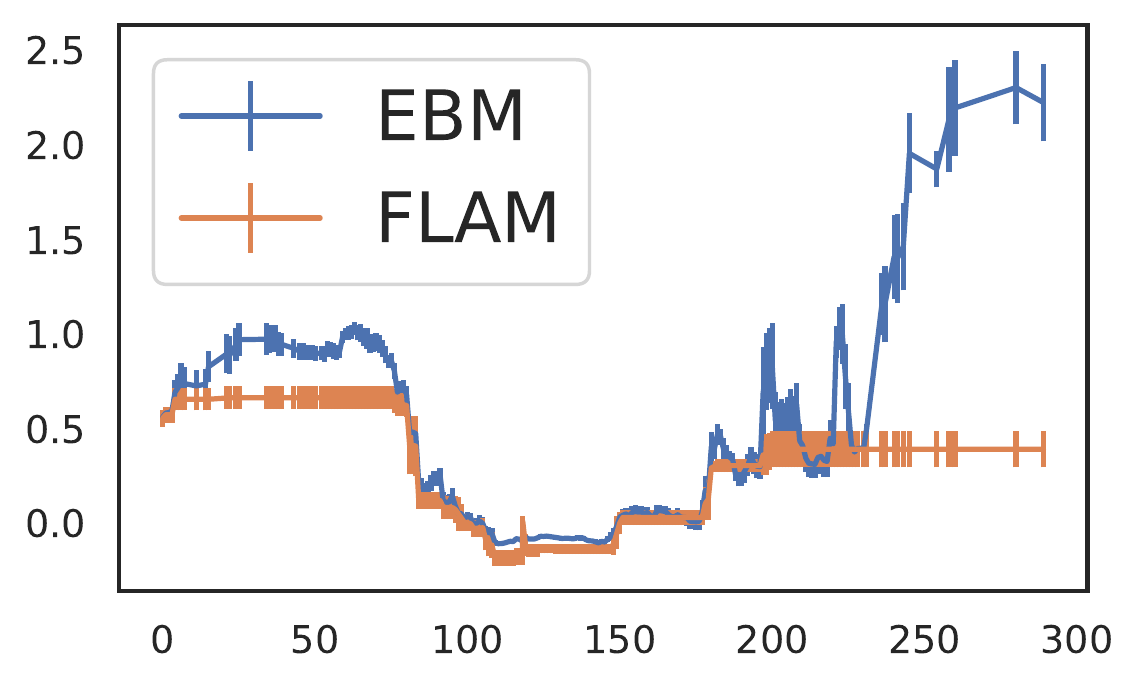}
 & \includegraphics[width=0.2\linewidth]{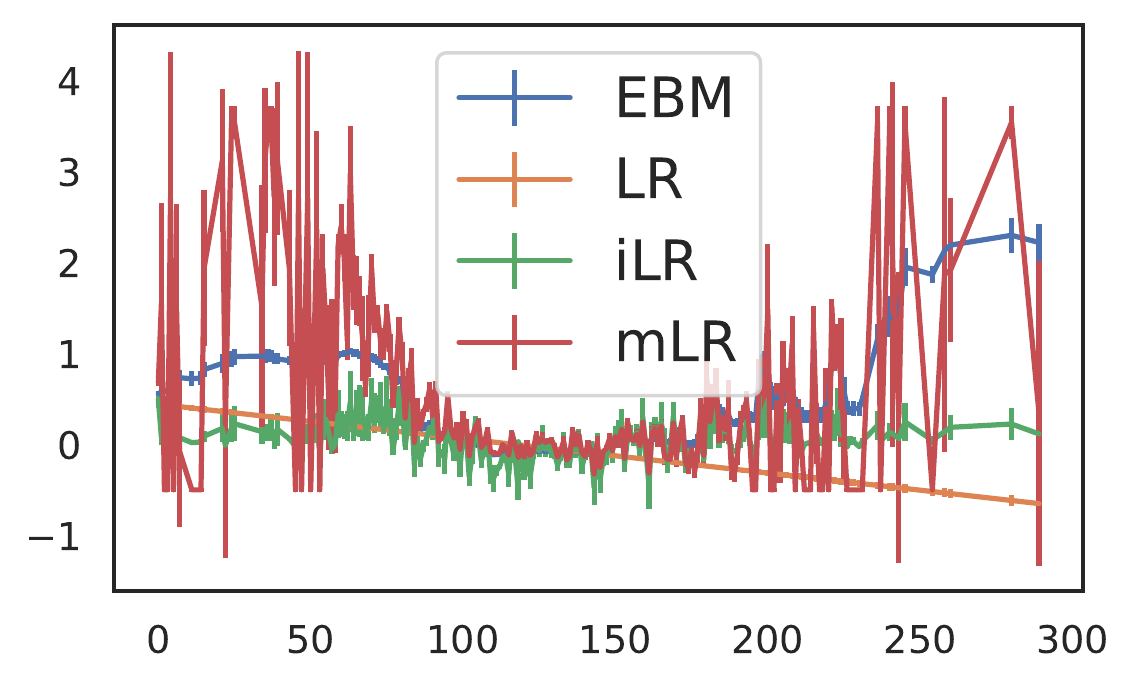} 
 \\
 
 & AIDS & AIDS & AIDS & AIDS & AIDS \\
 \raisebox{3.\normalbaselineskip}[0pt][0pt]{\rotatebox[origin=c]{90}{\small Log odds}} 
 & \includegraphics[width=0.2\linewidth]{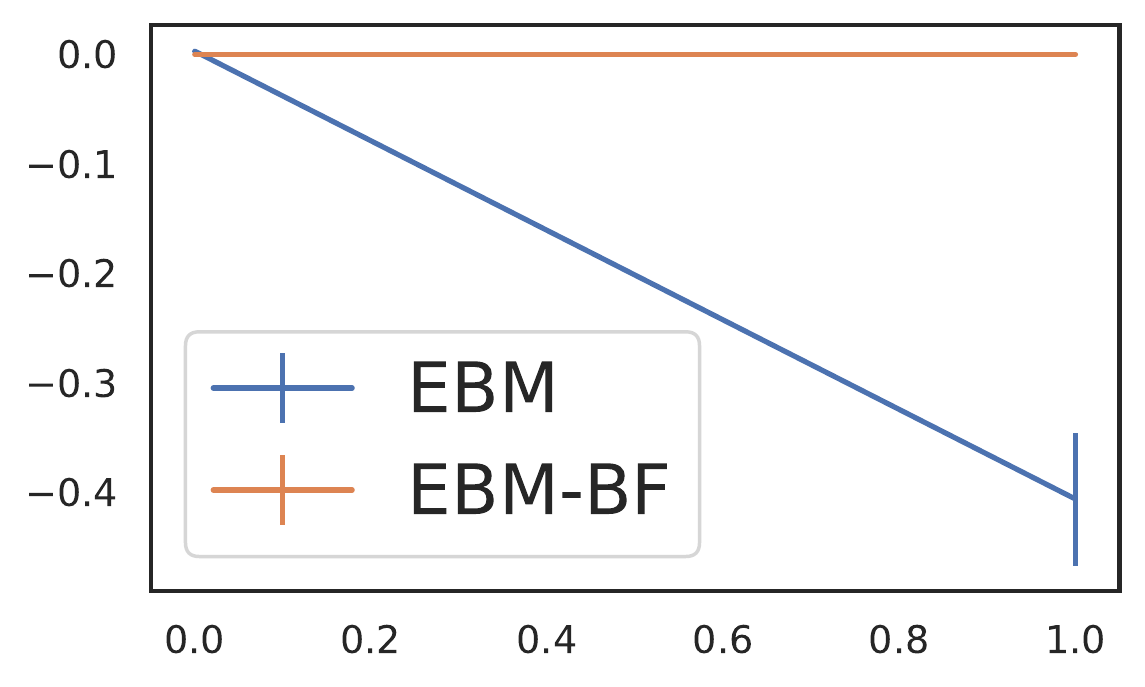}
 & \includegraphics[width=0.2\linewidth]{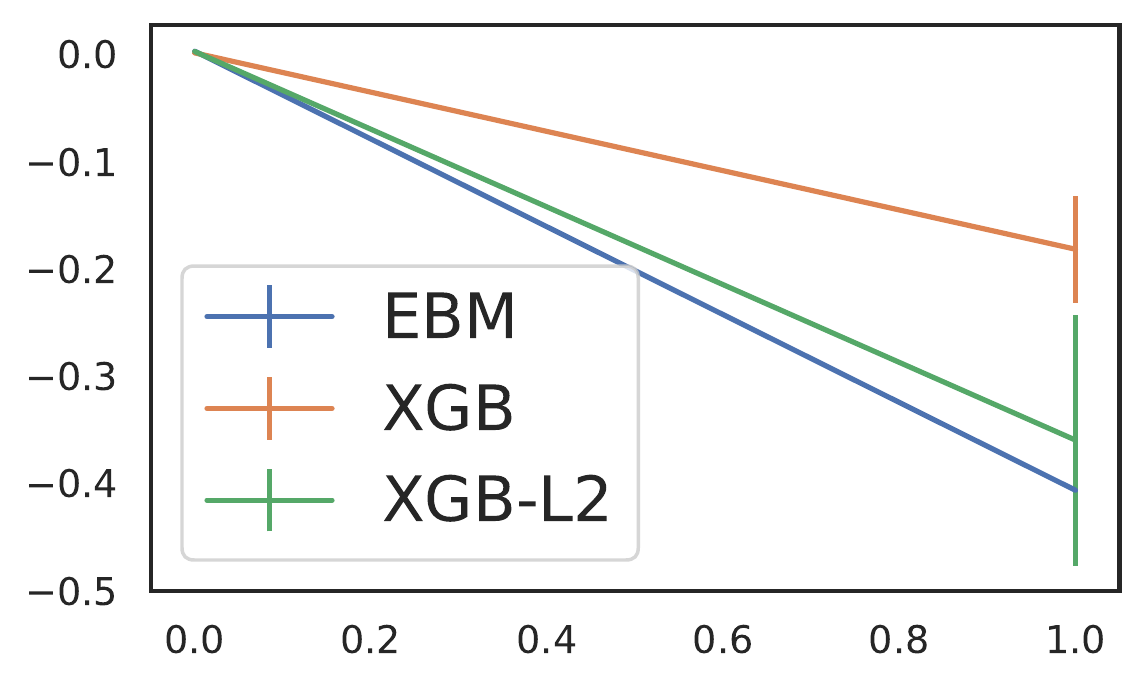}
 & \includegraphics[width=0.2\linewidth]{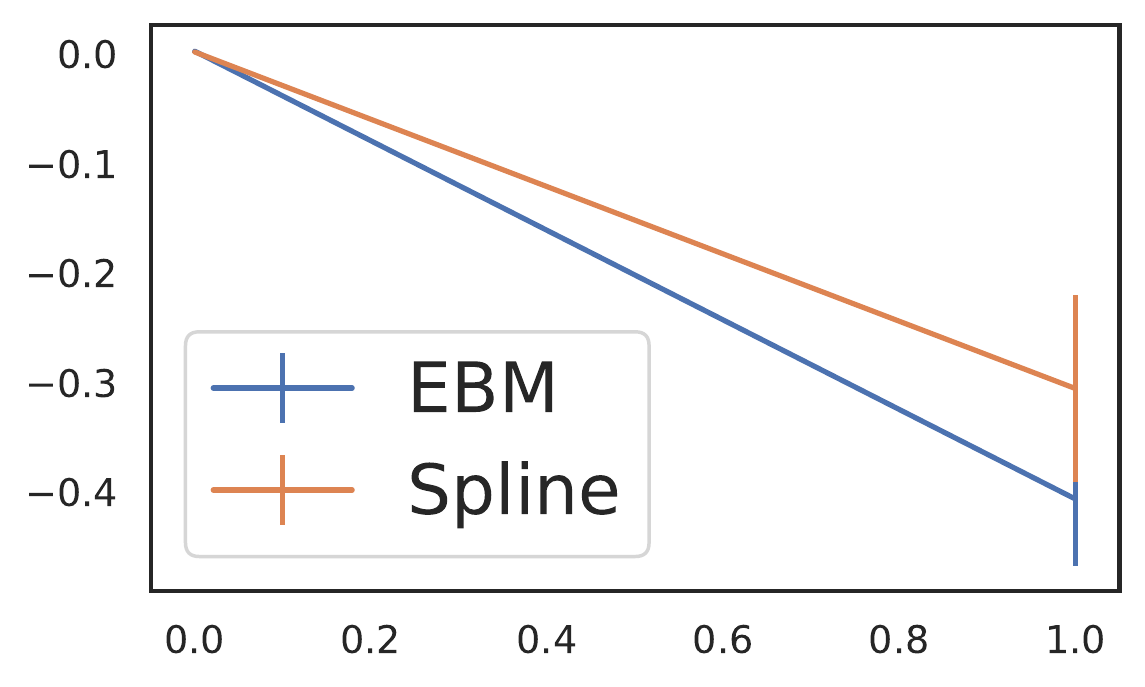}
 & \includegraphics[width=0.2\linewidth]{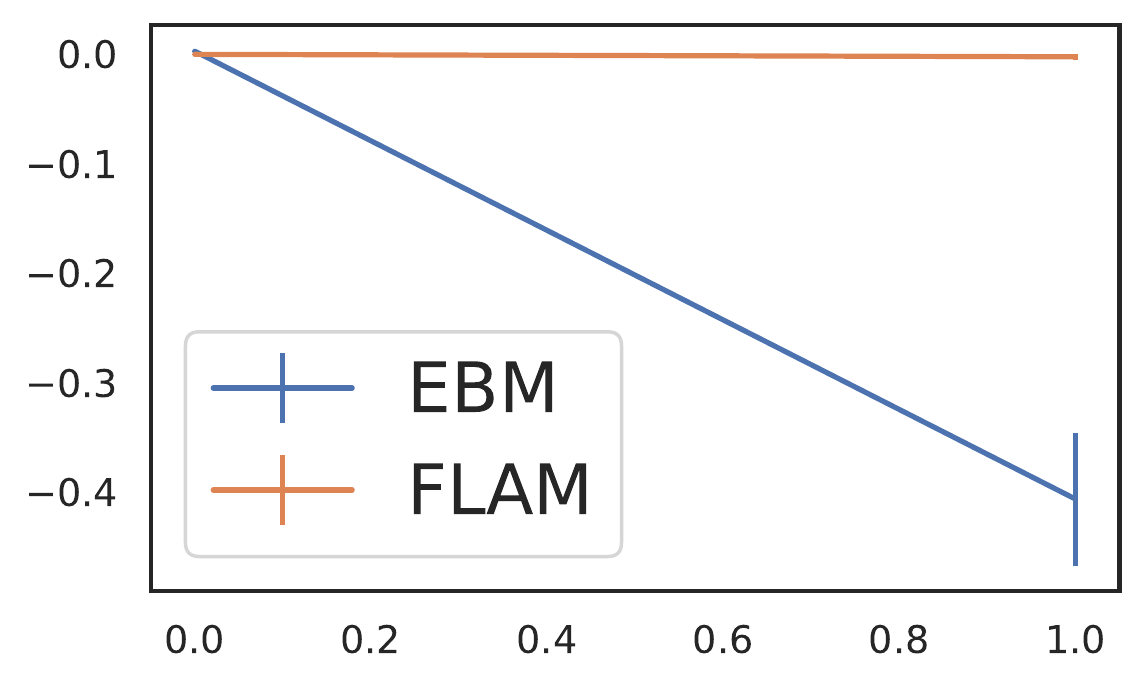}
 & \includegraphics[width=0.2\linewidth]{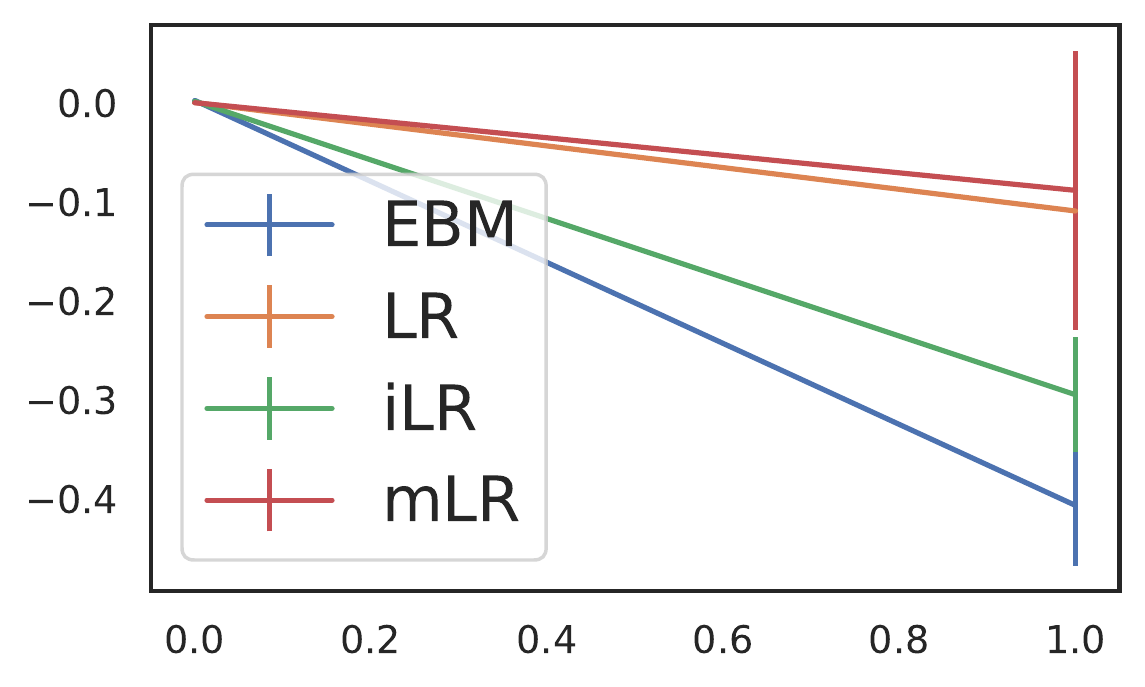} 
 \\
 
  \end{tabular} 
\end{center}

  \caption{
     Shape plots from nine GAM algorithms trained on the MIMIC-II dataset (three of seventeen features shown). To make comparisons easier EBM (blue line) is repeated in each plot.
  }
  \label{fig:fig1}
\end{figure*}

Do GAMs trained with different algorithms agree with each other? In Fig.~\ref{fig:poster_child}, we show that three GAMs with similar accuracy provide very different interpretations of the COMPAS recidivism dataset, a dataset in which bias is an important concern. For instance, in Fig.~\ref{fig:fig1}(a) EBM-BF suggests that there is no racial bias in the data, while Spline indicates that there is strong racial bias, yet is slightly less accurate. Should we believe EBM-BF because of its slightly higher accuracy and believe there is no racial bias? Probably not. Then how should we determine which GAM to believe?

In this paper, we try to answer this question by studying two aspects of GAMs trained with different algorithms. First, we quantify which GAMs use fewer features to make predictions (similar to $\ell_1$-regularization), which we call \emph{feature sparsity}. Although feature sparsity is sometimes preferred because it appears to yield simpler explanations~\citep{tibshirani1996regression, doshi2017towards}, it can be dangerous for data exploration as it can hide bias in the data. Consider a GAM that appears to be unbiased by showing no effect on sensitive variables such as race, but instead, because the learning algorithm is biased to use fewer features, it has compiled the racial bias into other correlated variables like zip code that are not obviously related to race, thus allowing the racial bias to go unrecognized.
Furthermore, for features that only matter for rare subpopulations (e.g. a rare disease), a sparse-feature GAM could easily ignore such features but still remain accurate, leading to failure or discrimination for that subpopulation. In this paper, we empirically verify this phenomenon by showing that sparse GAMs often have higher loss for minority classes than less sparse GAMs.

Second, we examine how much we can trust each GAM to reflect true patterns in the data, a property we call \emph{data fidelity}. \citet{shmueli2010explain} contrasted predictive models that seek to minimize the \textit{combination} of variance and bias (defined in a statistical sense, not in the sense of unfairness) to explanatory models that aim to capture true patterns in data by minimizing bias alone. For the former, bias can be sacrified for improved variance, and \citet{shmueli2010explain} provided examples of how the ``wrong'' model can sometimes predict better than the right one. In this paper, we study this phenomenon across different GAM algorithms. For real data where we do not know the underlying data patterns, we use the bias term from bias-variance analysis as a proxy for data fidelity. We also experiment with simulated datasets that have different data generators, each of which may favor GAM algorithms with certain inductive biases, and measure the worst-case data fidelity of each GAM algorithm across multiple datasets. This allows us to quantify if some GAM algorithms have high accuracy but low data fidelity, which may mislead users to trust the wrong explanations.

Our key contributions in this paper are:
\begin{itemize}[leftmargin=*]
    \item We compare different GAM algorithms on ten classification datasets and find that the most accurate GAMs yield similar accuracy, yet learn qualitatively different explanations.

    \item We measure which GAM algorithms lead to models that are more or less sparse, a property we call \textit{feature sparsity}. We show that sparse-feature GAMs can discriminate on rare subpopulations leading to unfairness.
    
    \item We examine several case studies of data anomaly discovery to see which GAMs can or cannot be trusted to discover these true patterns in the data, a property we call \textit{data fidelity}. 
    
    \item We show that some GAMs have high accuracy but low data fidelity which can mislead users who select models by accuracy alone.
    
    \item We find that inductive bias plays a crucial role in model explanations, and recommend tree-based GAMs over other GAMs for their low feature sparsity and superior data fidelity.

\end{itemize}

%    \item We introduce two new metrics: one to measure feature sparsity, and one to measure data fidelity.

% We compare the data fidelity of GAMs trained on real datasets by bias-variance analysis, and on simulated datasets generated with different inductive biases to see which GAMs have the best worst-case data fidelity.

\section{Related Work}
\label{sec:related_works}

% Several desirable properties have been proposed as necessary in order to call a model or explanation interpretable. These include stability, robustness, and insensitivity of explanations \cite{alvarezmelis2018robustness,alvarez2018towards,yeh2019infidelity,zhang2019trust, dombrowski2019explanations,adebayo2018sanity}. Underlying these ideas as well as our work is the belief that different, even contradictory explanations of the same model or data reduces the explanation's trustworthiness \cite{poursabzi2018manipulating}, hence reducing its interpretability. 
% Our finding that using test set accuracy to choose among interpretable models can be misleading is consistent with recent results on post-hoc explanations where it is possible to find similarly accurate explanations that are biased with regards to the underlying model they attempt to explain \cite{lakkaraju2020fool,ghorbani2019interpretation}. 

While we study GAMs in this paper, GAMs are not the only interpretable model class to come under scrutiny recently. The instability of decision trees (another model class commonly considered interpretable) has been pointed out \citep{dwyer2007instability}, and the vulnerability of post-hoc explanation methods such as LIME \citep{lime} and Shapley values \citep{lundberg2017unified} to input perturbation  has been exploited to generate adversarial attacks on model explanations \citep{slack2020fooling}. \citet{hooker2019permuting} also found partial dependence and feature importance metrics based on permuting inputs to be particularly misleading when inputs are highly dependent, and earlier work found feature importance metrics to be biased for certain types of models with different inductive biases, e.g., random forest feature importance is biased towards variables with many potential splits such as categorical variables with many levels \citep{strobl2007bias,zhou2019unbiased}. 

Our paper is not the first to compare different GAM algorithms, but to the best of our knowledge it is the first to focus on interpretability and its relationships to fairness on different GAM algorithms. \citet{binder2008comparison} compared three different spline training algorithms, including backfitting, joint optimization, and boosting, finding that boosting performed particularly well in high-dimensional settings. 
\citet{lou2012intelligible} also found that boosted shallow bagged trees yielded higher accuracy than other GAM algorithms. 
However both papers focused on accuracy, not interpretability.

\setlength\tabcolsep{0.1pt} % default value: 6pt

\section{Methods}
\label{sec:gam_intro}
In this section we describe the different GAM algorithms used in this paper. To make it easier for readers, we defer the description of the new metrics we define in this paper -- feature sparsity and data fidelity -- to just before their use in Sec.~\ref{sec:results}.

\subsection{GAM Algorithms}
Given an input $x\in\mathbb{R}^{N\times D}$, a label $y$, a link function $g$ (e.g. in binary classification, $g$ is logit), and shape functions $f_j$ for each feature, a generalized additive models (GAM) can be written as:
\begin{equation}
\label{eqn:gam}
    g(y) = f_0 + \sum_{j=1}^D f_j(x_j).
\end{equation}
GAMs are interpretable because the impact of each feature, $f_j$, on the prediction can be visualized as a graph (see Fig.~\ref{fig:fig1} for an example), and humans can easily simulate how a GAM works by reading $f_j$s off different features from the graph and adding them together.
We select the following six GAM algorithms to compare in this paper based on their popularity, state-of-the-art performance and availability of open source implementations. 

\textbf{Explainable Boosting Machine (EBM) \ \ } A tree-based GAM designed for intelligibility and high accuracy \citep{lou2012intelligible,caruana2015intelligible,nori2019interpretml} where shape functions $f_j$ are gradient-boosted ensembles of bagged trees. Each tree operates on a single variable, preventing interactions effects from being learned. Trees are grown by repeatedly cycling through features, which forces the model to sequentially consider each feature as an explanation of the current residual rather than greedily selecting the best feature. This deliberate construction makes this model have less \emph{feature sparsity}. 
For comparison, we create a sparse version of EBM similar to regular gradient boosted trees, "EBM-BF" (EBM-BestFirst), that greedily grows the next tree on the best, most informative feature to reduce error as much as possible at each step. Like most gradient boosted trees, EBM-BF is likely to put most weight on a few very important features, modest weight on a larger number of moderately useful features, and little or no weight on features whose signal could be learned by other stronger, correlated features.

\textbf{XGBoost (XGB) \ \ }
We introduce a new tree-based GAM based on the popular boosting package XGBoost~\citep{Chen:2016:XST:2939672.2939785}. 
% XGBoost improves upon regular boosted trees by loss derivatives and weight regularization to improve training speed and accuracy.
To convert XGB to a GAM, we limit tree depth to $1$ (stumps) so that the trees are not able to learn feature interactions, and we bag XGB to improve accuracy (similar to EBM).
% This means that each tree is only allowed to have depth $1$ and thus not able to 
% Then we bag XGB $100$ times to improve its accuracy (similar to EBM).
% We find that bagging XGB depth-$1$ models significantly improves their accuracy, and since a bagged GAM is still a GAM, we use $100$ bags for the model.  
% (similar to the finding with EBMs \citep{caruana2015intelligible}). But for XGB with depth $3$, bagging surprisingly hurts performance a bit, which might explain why it's uncommon to do bagging with boosted trees.
% We set bagging parameter to $100$ for XGB in the rest of the paper.
We also create a new version of XGB, "XGB-L2", similar to EBMs, that picks features sequentially when growing trees instead of greedily choosing the best feature.
To achieve this, we set the XGB random subsampling of features parameter to a small ratio such that each tree is given just $1$ feature.
This deliberate modification makes this model have less feature sparsity. 
This modification makes XGB more of a "dense" model similar to $\ell_2$ regularization that often uses all features.
Fig.~\ref{fig:fig1}(b) shows these $2$ methods.
To our surprise, although XGB and EBM are both boosted trees, their shape plots can be quite different (Fig.~\ref{fig:fig1}(b)).
% Shape plots learned by XGB, XGB-L2 and EBM tend to agree with each other in in large-sample regions, but diverge in  low-sample regions on Age, PFratio and SBP.

% XGB-L2 shows a pattern similar to EBM, while XGB is more similar to EBM-BF.
% On the Age shape plot, EBM has an interesting jump around $55$ while both XGB algorithms miss it.
% In SBP, EBM has higher peaks than XGB, especially at $x=200$ and $x=225$.  The jumps in risk that happen near SBP $175$, $200$, and $225$ are due to treatment effects. $175$, $200$, and $225$ are treatment thresholds that doctors use: as patient's SBP increases their risk naturally increases, but when they reach the next treatment threshold, risk actually drops because most patients just above the threshold are receiving more aggressive treatment that is effective at reducing their risk.
%In PFratio, all $3$ methods successfully capture the drop in $330$, since all the missing data is imputed by the mean value and missing data indicates lower risk.
%% Not true anymore though
% We also experimented with XGB with the subsampling parameter set to 0.5, yielding an intermediate point between extreme L1- and L2-ish behavior. This version of XGB yielded slightly higher accuracy than XGB, XGB-L2 and EBMs.

\textbf{Spline \ \ }
A classic way to train GAMs is with spline basis functions \citep{hastie1990generalized}. We tried a variety of spline methods in 2 popular packages, the Python pygam \citep{pygam} and R mgcv package \citep{mgcv}, and chose cubic splines in pygam because it has a good combination of accuracy, robustness and speed (Fig.~\ref{fig:fig1}(c)).

\textbf{Fused LASSO Additive Models (FLAM) \ \ }
For each unique value of feature $x_j$, Fused LASSO Additive Model (FLAM) \citep{petersen2016fused} learns a weight on each value, and adds an $\ell_1$ penalty to the differences between adjacent weights.
This $\ell_1$ penalty causes FLAM to produce relatively flat graphs and penalize unnecessary jumps.
We use the R package FLAM \citep{petersen2016fused} in our experiments (Fig.~\ref{fig:fig1}(d)).
% In Fig.~\ref{fig:fig1}(d), we show FLAM model is more similar to tree-based models but flatter and thus do not show lots of patterns. 

\textbf{Logistic regression (LR) and other strawmen approaches \ \ }
We compare these other approaches to Logistic Regression (\textbf{LR}), a widely used linear model that cannot learn non-linear shape plots. 
We also compare to two other strawmen approaches: marginalized LR (\textbf{mLR}) and indicator LR (\textbf{iLR}).
We first bin each feature $x_j$ into at most $255$ bins.
In contrast to LR that assumes $f_j(x_j) = w_jx_j$, \textbf{mLR} sets $f_j(x_j) = w_j g(x_j)$ where $g(x_j)$ is the average (marginalized) value of target $y$ within the same bin as $x_j$ in the dataset. This is a GAM model built by applying logistic regression on top of marginalization, thus preventing shape plots from being learned in concert with each other.
\textbf{iLR} treats each bin as a new feature (similar to one-hot encoding) and learns an LR on the transformed features. It thus ignores proximity relationships between different feature values (Fig.~\ref{fig:fig1}(e)). 
% As expected, the shape plots for LR are straight lines that miss much of the interesting detail in the data.  Despite this, LR has reasonably competitive accuracy in Table~\ref{table:real_data_test_AUC}.  
% iLR and mLR, however, are even less accurate than LR.  iLR appears to be too heavily regularized to learn the details in the MIMIC-II data, and mLR appears to not be regularized enough and shows high noise in Fig.~\ref{fig:fig1}(e).

\setlength\tabcolsep{0.pt} % default value: 6pt
\begin{figure*}[tbp]
  \begin{center}

\begin{tabular}{ccccc}
   & (a) COMPAS dataset -- race
   & (b) COMPAS dataset -- gender
   & (c) Adult dataset -- race
   & (d) Adult dataset -- gender \\
 \raisebox{-3.5\normalbaselineskip}[0pt][0pt]{\rotatebox[origin=c]{90}{\small Log odds}}
 & \imagetop{\includegraphics[width=0.25\linewidth]{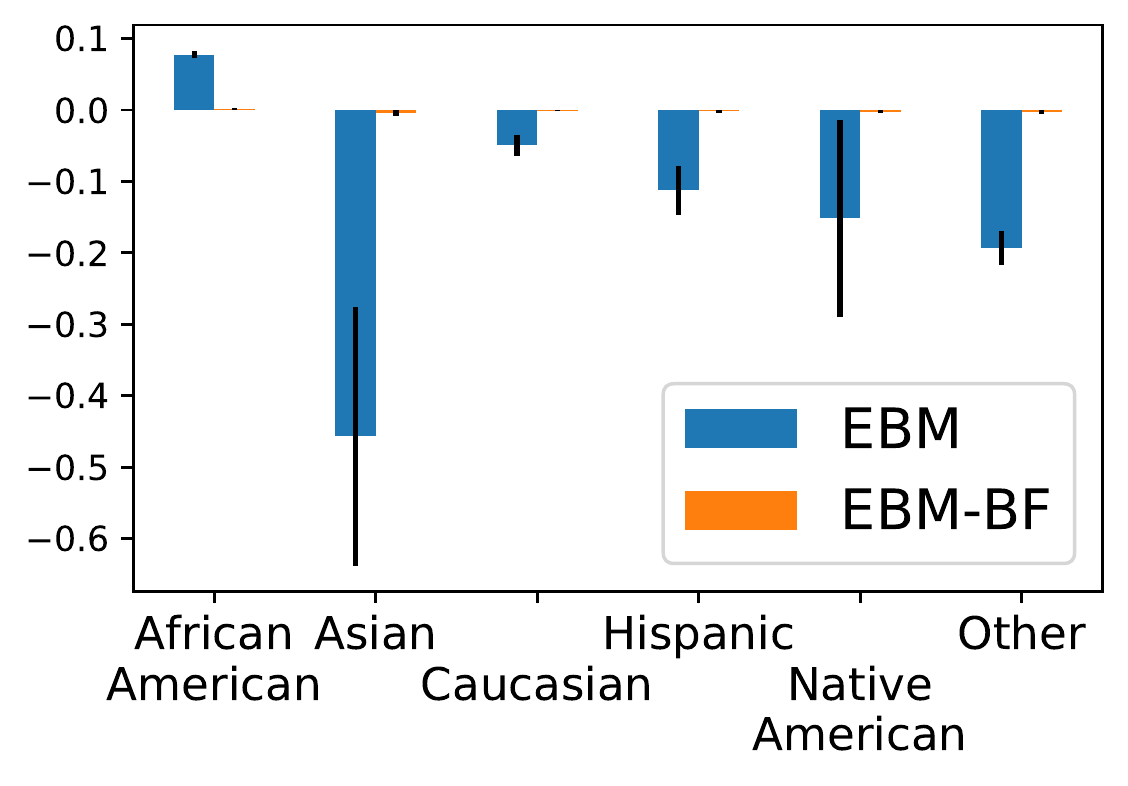}}
 & \imagetop{\includegraphics[width=0.25\linewidth]{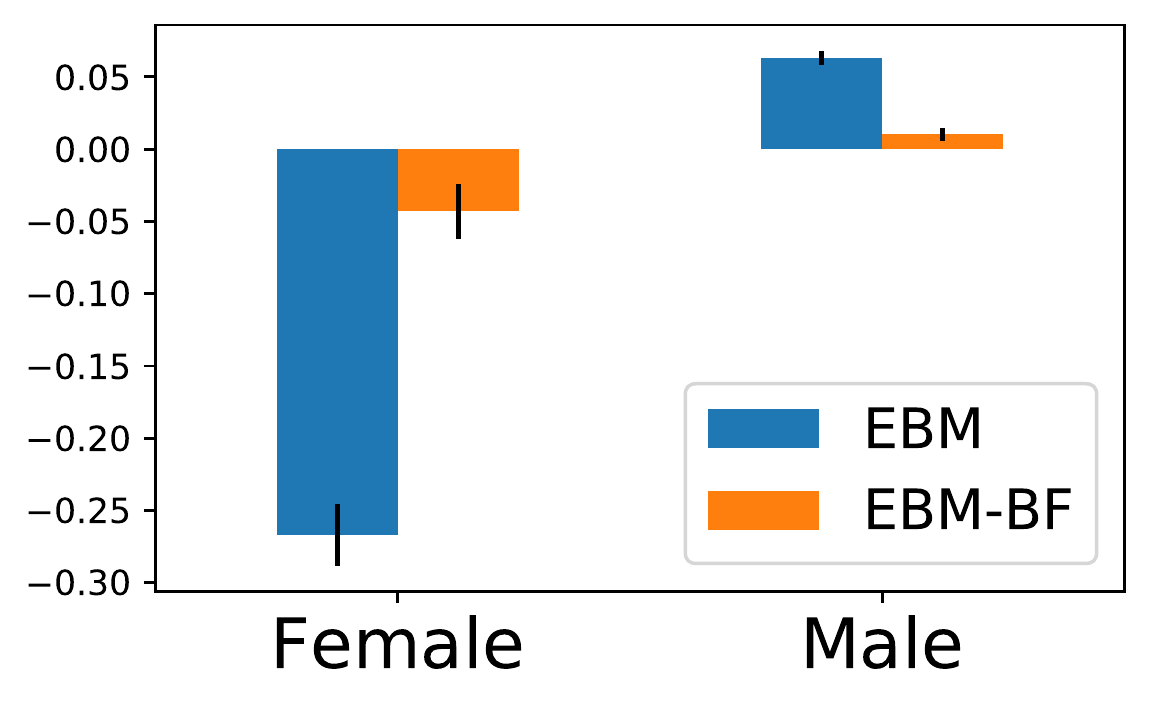}}
 & \imagetop{\includegraphics[width=0.25\linewidth]{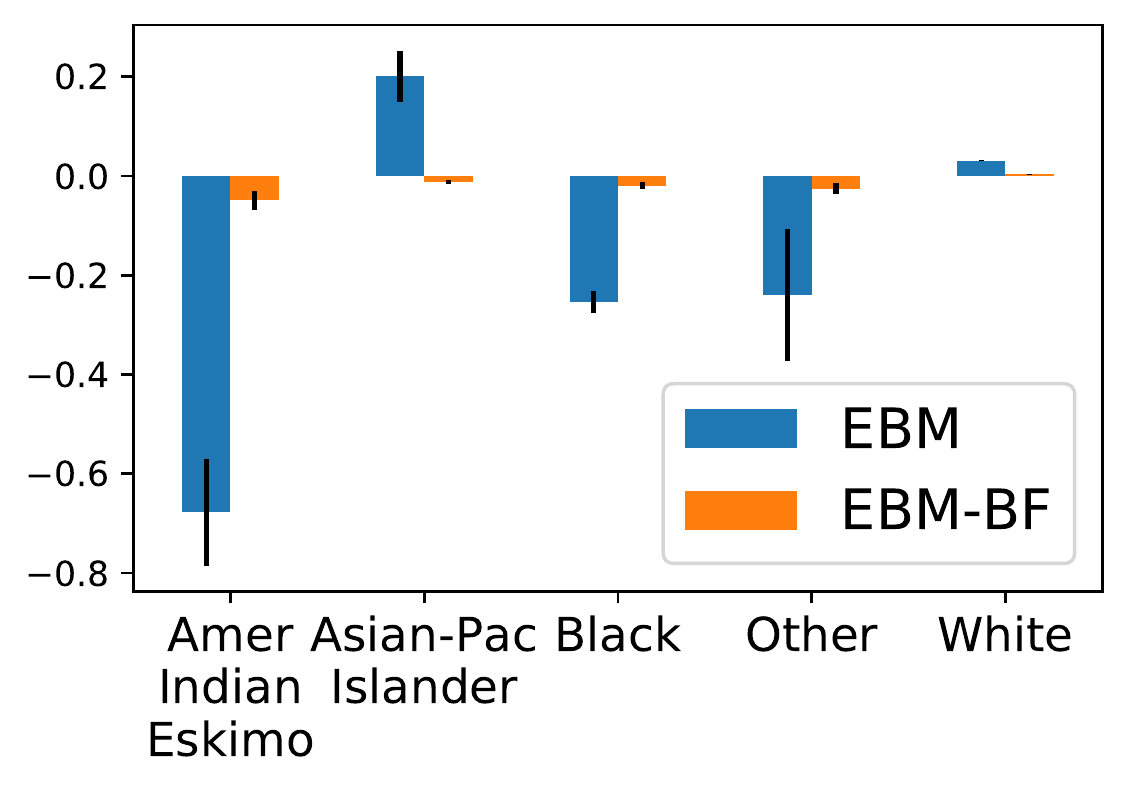}}
 & \imagetop{\includegraphics[width=0.25\linewidth]{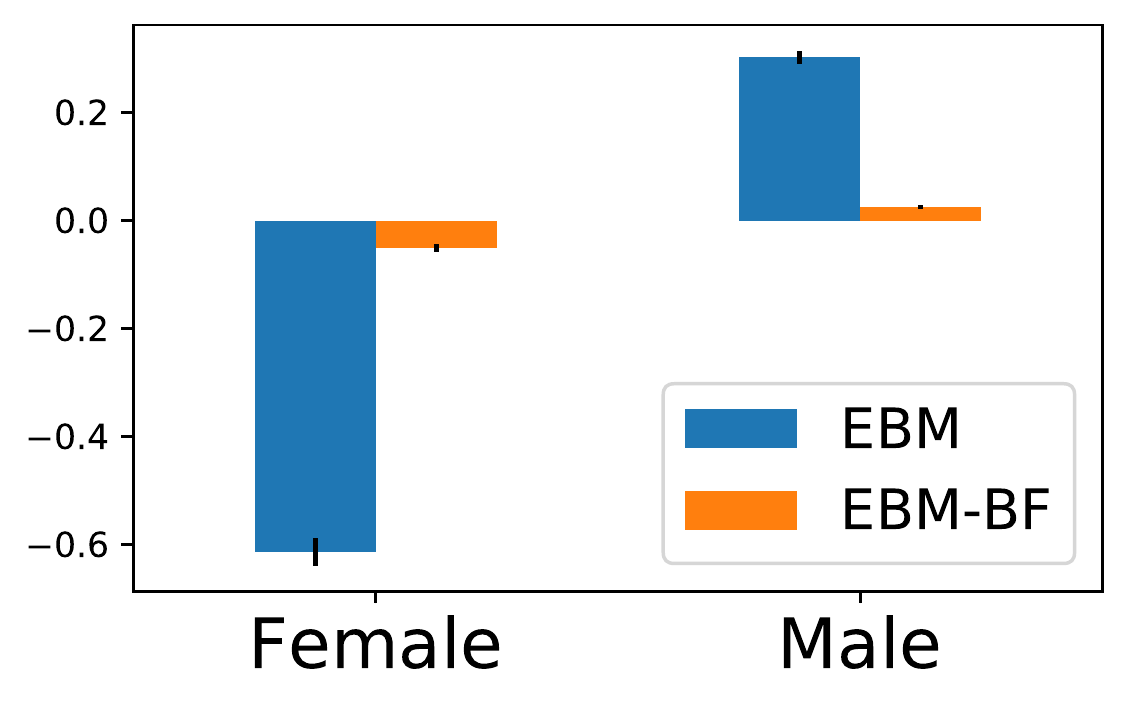}} \\
  \end{tabular}
    \end{center}
  \caption{
     Shape plots for COMPAS and Adult datasets for two sensitive attributes: race and gender. We compare two extreme GAMs: a dense-feature GAM (EBM) and a sparse-feature GAM (EBM-BF). Sparse EBM-BF learns little effect on these features.
  }
  \label{fig:fairness}
\end{figure*}

% \note{If need to cut length, combine this section into the Appendix and just leave a line "To fairly compare different GAM algorithms, we choose hyperparameters that perform best for each algorithm individually. We point the reader to the Appendix for details. We split each dataset into 70-15-15\% train-val-test splits and repeat our training procedure run five times. This allowed us to derive uncertainty estimates in the form of standard deviation across multiple runs. }

\subsection{Training and Hyperparameters \ \ }
To fairly compare different GAM algorithms, we choose hyperparameters that perform best for each algorithm individually. Below, we briefly mention how we tune each GAM algorithm, and point the reader to the Appendix for more details.

For tree-based methods \textbf{EBM} and \textbf{XGB}, we perform early stopping to determine the optimal number of trees, stopping when the validation set performance stops improving for more than $50$ trees. For \textbf{Spline}, we choose a maximum of $50$ knots and use the gcv criterion ~\citep{wahba1985comparison} to select the smoothness penalty. We found that using more than 50 knots is intractable for larger datasets and does not improve performance in smaller datasets. For \textbf{FLAM}, we cross-validate the $\lambda$ parameter and then refit the model on the entire training set using the optimal parameter. For \textbf{LR}, we cross-validate the $\ell_2$ penalization parameter.

We split each dataset into 70-15-15\% train-val-test splits and repeat our training procedure run $5$ times. This allowed us to derive uncertainty estimates in the form of standard deviation across multiple runs. 

%To derive uncertainty bars in the shape plots, we average the graph over the $5$ runs and plot the mean and standard deviation.

\section{Case studies: COMPAS, Adult, MIMIC-II}
\label{sec:real_world}
We start with some case studies to highlight the implications of different GAM algorithms on common interpretability tasks such as surfacing unfairness or discovering anomalies in data. In this section, we highlight our key findings with plots
specifically picked to be representative of our main results.
A complete set of plots can be found in Appendix~\ref{sec:appx_additional_shape_graphs}.

\setlength\tabcolsep{2pt} % default value: 6pt
\begin{table*}[tbp]
\caption{Cross entropy loss of GAMs on different subpopulations in the COMPAS dataset. n is the number of samples in the subpopulation. The percentage shown is relative to the performance of EBM. Columns are sorted by descending n.}
\label{table:fairness_compas}
\begin{tabular}{c|c|ccccc|cc}
 &
  \makecell{\textbf{All} \\ \textbf{(n=6172)}} &
  \makecell{\textbf{Black} \\ \textbf{(n=3175)}} &
  \makecell{\textbf{White} \\ \textbf{(n=2103)}} &
  \makecell{\textbf{Other} \\ \textbf{(n=343)}} &
  \makecell{\textbf{Asian} \\ \textbf{(n=31)}} &
  \makecell{\textbf{Native American} \\ \textbf{(n=11)}} &
  \makecell{\textbf{Male} \\ \textbf{(n=4997)}} &
  \multicolumn{1}{c}{\makecell{\textbf{Female} \\ \textbf{(n=1175)}}} \\ \toprule
EBM &
  $0.586$ &
  $0.591$ &
  $0.590$ &
  $0.542$ &
  $0.470$ &
  $0.571$ &
  $0.591$ &
  $0.564$ \\ \midrule
EBM-BF &
  \makecell{$0.589$ \\ $\bm{(0.49\%)}$} &
  \makecell{$0.595$ \\ $\bm{(0.72\%)}$} &
  \makecell{$0.590$ \\ $\bm{(-0.02\%)}$} &
  \makecell{$0.550$ \\ $\bm{(1.45\%)}$} &
  \makecell{$0.500$ \\ $\bm{(6.48\%)}$} &
  \makecell{$0.558$ \\ $\bm{(-2.26\%)}$} &
  \makecell{$0.594$ \\ $\bm{(0.41\%)}$} &
  \makecell{$0.569$ \\ $\bm{(0.82\%)}$} \\ \midrule
\makecell{EBM\\without race} &
  \makecell{$0.587$ \\ $\bm{(0.10\%)}$} &
  \makecell{$0.591$ \\ $\bm{(0.06\%)}$} &
  \makecell{$0.590$ \\ $\bm{(-0.01\%)}$} &
  \makecell{$0.544$ \\ $\bm{(0.31\%)}$} &
  \makecell{$0.498$ \\ $\bm{(6.08\%)}$} &
  \makecell{$0.579$ \\ $\bm{(1.39\%)}$} &
  \makecell{$0.593$ \\ $\bm{(0.18\%)}$} &
  \makecell{$0.563$ \\ $\bm{(-0.30\%)}$} \\ \midrule
\makecell{EBM\\without sex} &
  \makecell{$0.588$ \\ $\bm{(0.23\%)}$} &
  \makecell{$0.594$ \\ $\bm{(0.57\%)}$} &
  \makecell{$0.588$ \\ $\bm{(-0.23\%)}$} &
  \makecell{$0.547$ \\ $\bm{(0.95\%)}$} &
  \makecell{$0.464$ \\ $\bm{(-1.14\%)}$} &
  \makecell{$0.540$ \\ $\bm{(-5.32\%)}$} &
  \makecell{$0.592$ \\ $\bm{(0.06\%)}$} &
  \makecell{$0.570$ \\ $\bm{(0.99\%)}$}
\end{tabular}
\end{table*}

\setlength\tabcolsep{6pt} % default value: 6pt
\begin{table*}[tbp]
\caption{Cross entropy loss of GAMs on different subpopulations in the Adult dataset. 
% The percentage shown is relative to the performance of EBM.
}
\label{table:fairness_adult}
\begin{tabular}{c|c|ccccc|cc}
  &
  \makecell{\textbf{All} \\ \textbf{(n=32561)}} &
  \makecell{\textbf{White} \\ \textbf{(n=27816)}} &
  \makecell{\textbf{Black} \\ \textbf{(n=3124)}} &
  \makecell{\textbf{Asian/Pac} \\ \textbf{(n=1039)}} &
  \makecell{\textbf{Indian/Eskimo} \\ \textbf{(n=311)}} &
  \makecell{\textbf{Other} \\ \textbf{(n=271)}} &
  \makecell{\textbf{Male} \\ \textbf{(n=21790)}} &
  \makecell{\textbf{Female} \\ \textbf{(n=10771)}} \\ \toprule
EBM &
  $0.265$ &
  $0.277$ &
  $0.163$ &
  $0.309$ &
  $0.204$ &
  $0.137$ &
  $0.321$ &
  $0.152$ \\ \midrule
EBM-BF &
  \makecell{$0.279$ \\ $\bm{(5.27\%)}$} &
  \makecell{$0.291$ \\ $\bm{(5.17\%)}$} &
  \makecell{$0.171$ \\ $\bm{(5.37\%)}$} &
  \makecell{$0.326$ \\ $\bm{(5.61\%)}$} &
  \makecell{$0.219$ \\ $\bm{(7.13\%)}$} &
  \makecell{$0.164$ \\ $\bm{(19.05\%)}$} &
  \makecell{$0.336$ \\ $\bm{(4.78\%)}$} &
  \makecell{$0.163$ \\ $\bm{(7.39\%)}$} \\ \midrule
\makecell{EBM\\without race} &
  \makecell{$0.265$ \\ $\bm{(0.15\%)}$} &
  \makecell{$0.277$ \\ $\bm{(0.02\%)}$} &
  \makecell{$0.164$ \\ $\bm{(0.98\%)}$} &
  \makecell{$0.311$ \\ $\bm{(0.73\%)}$} &
  \makecell{$0.216$ \\ $\bm{(5.61\%)}$} &
  \makecell{$0.139$ \\ $\bm{(1.04\%)}$} &
  \makecell{$0.321$ \\ $\bm{(0.12\%)}$} &
  \makecell{$0.152$ \\ $\bm{(0.26\%)}$} \\ \midrule
\makecell{EBM\\without sex} &
  \makecell{$0.271$ \\ $\bm{(2.34\%)}$} &
  \makecell{$0.283$ \\ $\bm{(2.27\%)}$} &
  \makecell{$0.170$ \\ $\bm{(4.77\%)}$} &
  \makecell{$0.313$ \\ $\bm{(1.29\%)}$} &
  \makecell{$0.201$ \\ $\bm{(-1.62\%)}$} &
  \makecell{$0.138$ \\ $\bm{(0.47\%)}$} &
  \makecell{$0.326$ \\ $\bm{(1.54\%)}$} &
  \makecell{$0.161$ \\ $\bm{(5.78\%)}$}
\end{tabular}
\end{table*}

\subsection{How feature sparsity affects fairness?}
\label{sec:fairness}

% Consider a data set where there is significant correlation among features (which is very common).  
% A model using a few features ($\ell_1$-ness) will have used the correlation among features to “compile” the effect of weaker features into the stronger features, allowing it to place little or no learned effect on the weak features.
% But the L2 model will show potentially interesting effects on all or most of the features. 

% A model trained with an L1 bias will try to use as few features as possible, while a model trained with an L2 bias will be more likely to spread learned effects among many or all of the features.  
% Once again, imagine that the L1 and L2 models make similar predictions, i.e., one is not significantly more accurate than the other and there is a strong correlation between their predictions on test data (this, too, is common).  
% The L1 model will have used the correlation among features to “compile” the effect of weaker features into the stronger features that it retains, allowing it to place little or no learned effect on the weak features the L1 bias regularizes away.  

One key property we study in this paper is which GAM algorithm uses fewer features to make predictions i.e. \emph{feature sparsity}.
Although sparsity is sometimes preferred because it appears to generate simpler explanations, it can hide data bias and discriminate against minority groups.
Here we examine the sparsity properties of different GAM algorithms on two datasets that have been studied in the fairness community for racial and gender bias~\citep{zemel2013learning, chouldechova2017fair, mehrabi2019survey}. The COMPAS dataset contains demographic, crime, and recidivism information for defendants in Broward County, Florida, in $2013$ and $2014$. Research has suggested that the COMPAS recidivism risk score may be racially biased \cite{propublica}. The Adult dataset extracted demographic information, including age, race, occupation, sex, etc. from the 1994 census data to predict if an individual's income exceeds 50k/yr. In the dataset, males have on average higher annual incomes than females~\citep{mehrabi2019survey}.

To motivate our analysis, we compare two GAM algorithms that are very different from each other in terms of feature sparsity: sparse EBM-BF and regular, ``dense'' EBM, yet achieve similar accuracy (see Table \ref{table:real_data_test_AUC}). Figure \ref{fig:fairness} displays the shape plots on two sensitive attributes, race and gender, on the COMPAS dataset.
Since these features have modest influence compared to other features, the sparse-feature EBM-BF shows no or only a tiny effect on these sensitive attributes, while EBM shows much larger effects.
Although there is no easy way to judge which GAM is more ``causally" correct, the sparse EBM-BF makes users unaware of bias that may exist in the data and has been learned by other stronger, correlated features.
In contrast, the dense EBM shows effects on all features.
%, allowing humans to have a final say of whether or not to believe the patterns.
Because of this, we suggest that the dense model is better suited for surfacing potential bias in data then can then be investigated further by humans. 

% Consider a data set where there is significant correlation among features (which is very common). A model using a few features will have used the correlation to “compile” the effect of weaker features into the stronger features, allowing it to place little or no learned effect on the weak ones.
% It makes users unaware of such effect exists.
% In contrast, a dense model will show potentially interesting effects on all or most of the features, allowing human to have a final say of whether or not to believe the patterns.
% Therefore we argue dense model is more favorable in data bias discovery.
% this shows that the sparse nature of the model tends to ignore the non-important features and could easily hide the bias of the sensitive attributes.

Next, we investigate how feature sparsity affects minority groups. Table \ref{table:fairness_compas} presents the predictive performance (cross entropy loss) of EBM and EBM-BF on each minority group.
Although EBM and EBM-BF have negligible difference (less than $0.5\%$) in terms of overall loss, compared to EBM, EBM-BF exhibits greater loss on minority groups Other ($1.45\%$) and Asian ($6 \%$) compared to majority group White ($-0.02\%$); EBM exhibits lower loss on the Native American group ($-2.26 \%$). To further investigate this phenomenon, we perform an ablation study by removing the race feature from EBM thus forcing EBM to be more sparse. While this increased overall loss by $0.1\%$ compared to EBM with the race feature, the loss for minority groups was again substantially increased, with the loss increasing by $6\%$ for Asian and $1\%$ for Native American. Similarly, when we remove the sex feature from EBM, the loss for the minority group Female increased by $0.99\%$, almost four times larger than the overall loss increase ($0.23\%$).
Unexpectedly, removing the sex feature improves the loss for minority group Native Americans ($-5.32\%$); this is a possible explanation for why the loss for Native Americans is smaller for EBM-BF than EBM, as EBM-BF placed little importance on sex. 

We repeat the same analysis on the Adult dataset. Table~\ref{table:fairness_adult} presents the loss of EBM and EBM-BF on each minority group in the Adult dataset. Compared to EBM, EBM-BF exhibit greater loss on minority groups Indian ($7.13\%$) and Other ($19.05\%$), much more than the overall loss ($5.27\%$) or loss on majority group White ($5.17\%$). 
We also find that removing race from the EBM model increased the loss more for minority groups Indian ($5.61\%$) and Other ($1.04\%$), and removing sex from the EBM model increases the loss for Female ($5.78\%$) much more than for Male ($1.54\%$).

\textbf{Implications \ \ }
GAM algorithms with a tendency to use fewer features to make predictions (e.g. EBM-BF) showed only small effects on sensitive attributes and exhibited greater prediction loss on minority groups causing unfairness, compared to GAM algorithms that tend to use more features to make predictions (e.g. EBM). 

%A sparse-feature GAM can miss important patterns in the data. so we suggest that dense GAMs are preferable for bias discovery. 
%Also, sparse GAMs are more likely to harm minority groups by ignoring weaker features that might be important for those groups.
% but not for majority groups.

\setlength\tabcolsep{0.pt} % default value: 6pt
\begin{figure}[tbp]
  \begin{center}

\begin{tabular}{ccc}
   & (a) PFratio
   & (b) Systolic BP
   \\
   \raisebox{3.\normalbaselineskip}[0pt][0pt]{\rotatebox[origin=c]{90}{\small Log odds}}
   & \includegraphics[width=0.5\linewidth]{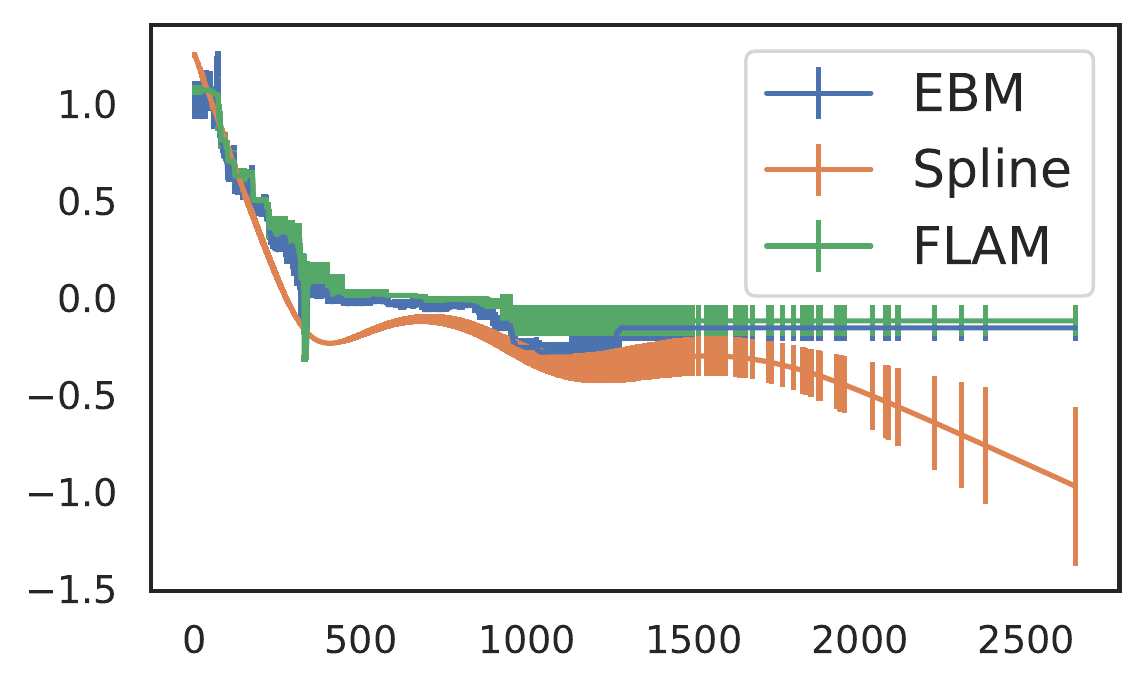}
   & \includegraphics[width=0.5\linewidth]{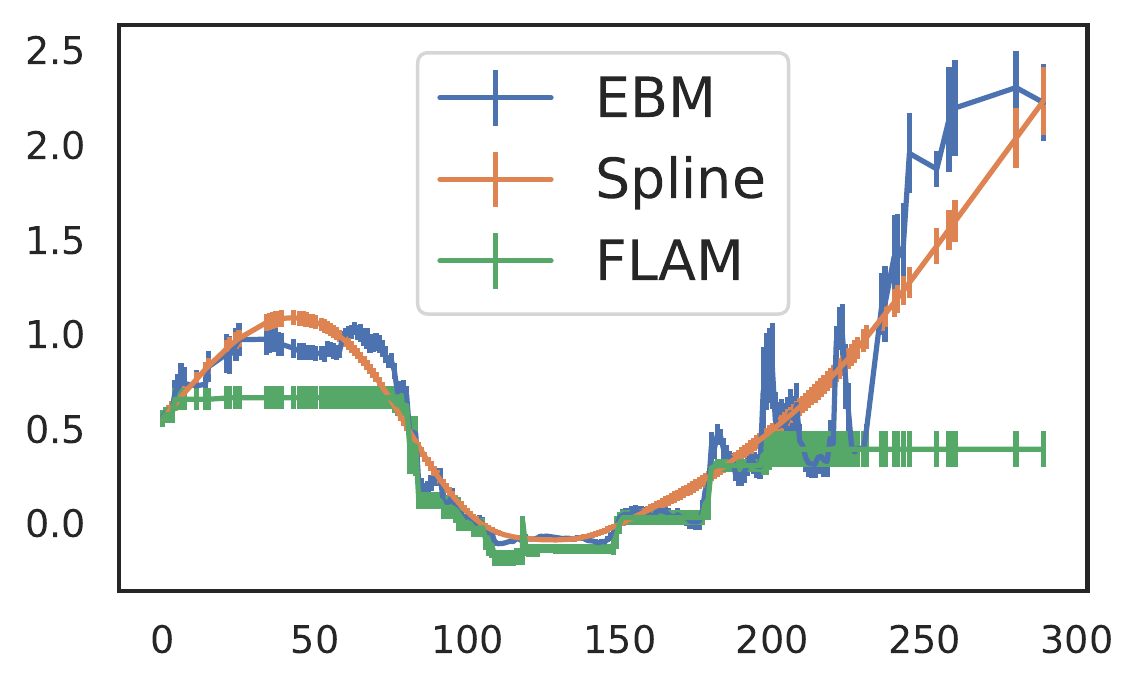} \\
  \end{tabular} 
\end{center}

  \caption{
      Two data anomalies in MIMIC-II that can be detected by tree-based GAMs (EBM):
      (a) PFratio missing values imputed using population mean $332$; (b) Systolic BP with likely human intervention artifacts at $175$, $200$ and $225$.}
  \label{fig:mimicii_case_study}
\end{figure}

% \note{Sarah: Needs a retitle. Maybe something like "How different GAMs have varying abilities to detect anomalies in data"?.}

\subsection{Data Anomaly Discovery}
\label{sec:anomaly}
Another key property we study in this paper is which GAM algorithm is better able to capture anomalies in data. 
To illustrate, we train different GAM algorithms on a medical dataset: ICU mortality prediction dataset MIMIC-II~\citep{johnson2016mimic}. On this dataset, XGB and EBM have similar shape plots thus we only present the EBM plots here for simplicity. 

% Here we demonstrate a dataset anomaly in a hopistal ICU mortality prediction task MIMIC-II~\citep{johnson2016mimic} that is caused by mean imputation and could be easily detected by tree-based GAMs (EBM, XGB) or FLAM, but not Spline.

Fig.~\ref{fig:mimicii_case_study}(a) displays one feature, PFratio (a measure of how well patients convert oxygen in air to oxygen in blood), for the three most accurate GAM algorithms on this dataset: EBM, Spline and FLAM.
Interestingly, both EBM and FLAM capture a sharp drop in mortality risk at PFratio=$332$.
It turns out that PFratio is usually not measured for healthier patients, and the missing values for these patients have been imputed by its population mean $332$ (a common preprocessing fix for missing data), thus giving a group of low-risk patients the mean value of this feature.
However, Spline is unable to represent the sharp drop, becoming distorted in the region 300-600, thereby underestimating the risk for patients in this region.

Fig.~\ref{fig:mimicii_case_study}(b) for Systolic Blood Pressure (BP) shows another data anomaly that is only captured by tree-based GAM algorithms EBM and XGB. EBM captures three jumps, exhibiting dips in risk predictions near $175$, $200$ and $225$. These are likely to be human intervention artifacts, since $175$, $200$, and $225$ are treatment thresholds used by physicians. As a patient's Systolic BP increases the mortality risk naturally increases, but when they reach the next treatment threshold, risk actually drops because most patients just above the threshold are receiving more aggressive care that is effective at reducing their risk.
Both Spline and FLAM are too smooth or flat and fail to capture these anomalies.

\textbf{Implications \ \ }
Localized data anomalies such as mean imputation and human intervention artifacts (e.g. medical treatment thresholds), often require models to learn quick, non-linear changes in risk. Tree-based methods (e.g. EBM and XGB) can detect these much better compared to GAM algorithms that are too smooth or sparse (e.g. Spline and FLAM).

\setlength\tabcolsep{1.1pt} % default value: 6pt
\begin{table}[t]

\centering
% \begin{small}
\caption{Description of ten classification datasets used.}

\begin{tabular}{c|cccc|c}
 & Domain & N & P & Positive Rate & Description \\ \toprule
Adult & Finance & 32,561 & 14 & 24.08\% & Income prediction \\
Breast & Healthcare & 569 & 30 & 62.74\% & Cancer classification \\
Churn & Retail & 7,043 & 19 & 26.54\% & Subscription churner \\
Credit & Retail & 284,807 & 30 & 0.17\% & Fraud detection \\
COMPAS & Law & 6,172 & 6 & 45.51\% & Reoffense risk scores \\
Heart & Healthcare & 457 & 11 & 45.95\% & Heart Disease \\
MIMIC-II & Healthcare & 24,508 & 17 & 12.25\% & ICU mortality \\
MIMIC-III & Healthcare & 27,348 & 57 & 9.84\% & ICU mortality \\
Pneumonia & Healthcare & 14,199 & 46 & 10.86\% & Mortality \\
Support2 & Healthcare & 9,105 & 29 & 25.92\% & Hospital mortality
\label{table:dataset_statistics}

\end{tabular}
% \end{small}

\end{table}

\section{Quantitative Analysis of GAMs}
\label{sec:results}

In the previous section, we saw examples of how different GAM algorithms revealed different insights. In this section, we study the performance differences between GAM algorithms quantitatively.
We first benchmark the test accuracy of different GAMs on ten different datasets (Sec.~\ref{sec:dataset}).
Then we measure feature sparsity of different GAM algorithms (Sec.~\ref{sec:l2_ness}). Finally, we measure data fidelity using both real (Sec.~\ref{sec:bias_var}) and simulated data (Sec.~\ref{sec:ss_graph_fidelity}, \ref{sec:ss_correlation_btw_explanation_and_generalization}).

\subsection{GAM accuracy}
\label{sec:dataset}

\setlength\tabcolsep{2pt} % default value: 6pt
\begin{table*}[tb]
\centering
\caption{Test set AUCs ($\%$) across ten datasets average over five runs. Best number in each row is in \textbf{bold}.}
\begin{tabular}{cccccccccccc}
 & \multicolumn{9}{c}{GAM} & \multicolumn{2}{c}{Full Complexity} \\
\cmidrule(lr){2-10}\cmidrule(lr){11-12}
 & EBM & EBM-BF & XGB & XGB-L2 & FLAM & Spline & iLR & LR & mLR & RF & XGB-d3 \\ \toprule
\multicolumn{1}{c|}{Adult} & $\bm{93.0} \pm 0.5$ & $92.8 \pm 0.5$ & $92.8 \pm 0.6$ & $91.7 \pm 0.6$ & $92.5 \pm 0.6$ & $92.0 \pm 0.6$ & $92.7 \pm 0.5$ & $90.9 \pm 0.6$ & $92.5 \pm 0.4$ & $91.2 \pm 0.5$ & $\bm{93.0} \pm 0.6$ \\
\multicolumn{1}{c|}{Breast} & $99.7 \pm 0.5$ & $99.5 \pm 0.5$ & $99.7 \pm 0.5$ & $99.7 \pm 0.5$ & $\bm{99.8} \pm 0.3$ & $98.9 \pm 0.8$ & $98.1 \pm 0.5$ & $99.7 \pm 0.4$ & $98.5 \pm 0.5$ & $99.3 \pm 1.1$ & $99.3 \pm 1.1$ \\
\multicolumn{1}{c|}{Churn} & $\bm{84.4} \pm 0.7$ & $84.0 \pm 0.9$ & $84.3 \pm 0.7$ & $84.3 \pm 0.7$ & $84.2 \pm 0.7$ & $\bm{84.4} \pm 0.8$ & $83.4 \pm 1.0$ & $84.3 \pm 0.7$ & $82.7 \pm 1.0$ & $82.1 \pm 0.6$ & $84.3 \pm 0.7$ \\
\multicolumn{1}{c|}{COMPAS} & $74.3 \pm 1.4$ & $\bm{74.5} \pm 1.7$ & $\bm{74.5} \pm 1.5$ & $74.3 \pm 1.5$ & $74.2 \pm 1.7$ & $74.3 \pm 1.5$ & $73.5 \pm 1.3$ & $72.7 \pm 1.0$ & $72.2 \pm 1.3$ & $67.4 \pm 1.2$ & $\bm{74.5} \pm 1.5$ \\
\multicolumn{1}{c|}{Credit} & $98.0 \pm 0.5$ & $97.3 \pm 1.3$ & $98.0 \pm 0.6$ & $98.1 \pm 0.6$ & $96.9 \pm 0.4$ & $\bm{98.2} \pm 0.7$ & $95.6 \pm 0.6$ & $96.4 \pm 1.1$ & $94.0 \pm 1.4$ & $96.2 \pm 1.5$ & $97.3 \pm 0.7$ \\
\multicolumn{1}{c|}{Heart} & $85.5 \pm 6.9$ & $83.8 \pm 6.0$ & $85.3 \pm 6.3$ & $85.8 \pm 7.0$ & $85.6 \pm 6.7$ & $86.7 \pm 6.3$ & $85.9 \pm 6.3$ & $\bm{86.9} \pm 5.8$ & $74.4 \pm 5.3$ & $85.4 \pm 6.5$ & $84.3 \pm 4.6$ \\
\multicolumn{1}{c|}{MIMIC-II} & $83.4 \pm 0.9$ & $83.3 \pm 0.8$ & $83.5 \pm 1.0$ & $83.4 \pm 0.9$ & $83.4 \pm 1.0$ & $82.8 \pm 0.8$ & $81.1 \pm 1.0$ & $79.3 \pm 0.8$ & $81.6 \pm 0.7$ & $\bm{86.0} \pm 0.6$ & $84.7 \pm 0.7$ \\
\multicolumn{1}{c|}{MIMIC-III} & $81.2 \pm 0.4$ & $80.7 \pm 0.7$ & $\bm{81.5} \pm 0.5$ & $\bm{81.5} \pm 0.5$ & $81.2 \pm 0.4$ & $81.4 \pm 0.4$ & $77.4 \pm 1.0$ & $78.5 \pm 0.5$ & $77.6 \pm 0.3$ & $80.7 \pm 0.8$ & $82.0 \pm 0.7$ \\
\multicolumn{1}{c|}{Pneumonia} & $\bm{85.3} \pm 0.6$ & $84.7 \pm 0.7$ & $85.0 \pm 0.8$ & $85.0 \pm 0.6$ & $\bm{85.3} \pm 0.9$ & $85.2 \pm 0.6$ & $84.3 \pm 1.0$ & $83.7 \pm 0.6$ & $84.5 \pm 0.7$ & $84.5 \pm 0.5$ & $84.8 \pm 0.8$ \\
\multicolumn{1}{c|}{Support2} & $81.3 \pm 1.0$ & $81.2 \pm 1.0$ & $81.4 \pm 1.1$ & $81.2 \pm 1.0$ & $81.2 \pm 1.1$ & $81.2 \pm 1.1$ & $80.0 \pm 1.2$ & $80.3 \pm 0.7$ & $77.2 \pm 0.9$ & $\bm{82.4} \pm 1.0$ & $82.0 \pm 1.4$ \\ \bottomrule
\multicolumn{1}{c|}{Average AUC} & $\bm{86.6}$ & $86.2$ & $\bm{86.6}$ & $86.5$ & $86.4$ & $86.5$ & $85.2$ & $85.3$ & $83.5$ & $85.5$ & $\bm{86.6}$ \\
\multicolumn{1}{c|}{Average Rank} & $3.70$ & $6.70$ & $\bm{3.40}$ & $4.90$ & $5.05$ & $4.60$ & $8.70$ & $7.75$ & $9.70$ & $7.40$ & $4.10$ \\
\multicolumn{1}{c|}{Normalized AUC} & $\bm{89.3}$ & $78.1$ & $87.3$ & $81.8$ & $83.6$ & $81.0$ & $47.4$ & $50.7$ & $28.5$ & $54.3$ & $86.5$
\label{table:real_data_test_AUC}
\end{tabular}
\end{table*}

How do we choose which GAM to use? Accuracy is perhaps the first obvious consideration. Table~\ref{table:real_data_test_AUC} provides test set AUC of different GAM algorithms on ten datasets. These datasets of varying size (500 - 250k samples) and number of features ($6$ - $57$ features) span different domains such as healthcare, criminal justice, finance, and retail (see Table \ref{table:dataset_statistics}). In addition to the nine GAM algorithms described in Sec.~\ref{sec:gam_intro}, we also include two full-complexity methods: Random Forest (RF) and XGB with depth $3$ (XGB-d3).
% Each dataset was randomly split into 70-15-15\% train-val-test splits and run for $5$ times. See Table \ref{table:appx_realdata_error_with_std} for results with standard deviation. 
% \note{Maybe remove Rank and Score to simplify story?}
For each method, we compute three metrics, each of which is averaged over ten datasets: (1) Test AUC; (2) Rank of test AUC compared to other methods (lower rank is better); (3) Test AUC normalized compared to other methods (lowest test AUC for a dataset has value 0, highest test AUC for a dataset has value 1, with all other test AUCs scaled linearly between them). On average across ten datasets, EBM, EBM-BF, and XGB-d3 performed the best. In general, GAMs perform better than or comparably to full complexity models. Four of the GAMs (EBM, XGB, Spline and FLAM) achieve similar top performance with average AUC differences less than $0.2\%$. 

% Among GAM methods, EBM and XGB perform the best, with XGB-L2, Spline and FLAM very close. 
% EBM-BF comes in fifth place, with the $3$ baselines, iLR, LR and mLR, performing the worst.
% For full complexity methods, XGB-d3 is better on average than RF, but performs slightly worse than EBM and XGB on two other metrics: Rank and Score.

\textbf{Implications \ \ } There exist GAM algorithms that perform comparably to full complexity models. Several GAM algorithms are similarly accurate, hence accuracy should not be the sole consideration when selecting between different GAM algorithms.

\subsection{GAM feature sparsity} 
\label{sec:l2_ness}

\setlength\tabcolsep{1.2pt}
\begin{table}[tbp]
\centering

\caption{Normalized feature density ($\%$). Higher numbers mean the model uses more features. Highest number in each row is in \textcolor{red}{red}; lowest number in each row is in \textcolor{blue}{blue}. Columns are sorted by descending density.}
\begin{tabular}{ccccccccccc}
  & XGB-L2  & EBM     & LR      & Spline  & XGB     & LASSO   & FLAM    & EBM-BF  \\ \toprule
Adult     & \color{red}{$33.9$} & $27.1$ & $22.0$ & \color{blue}{$20.5$} & $29.0$ & $21.3$ & $21.1$ & $22.6$ \\
Breast    & $11.2$ & $08.6$ & $13.0$ & \color{red}{$23.4$} & $07.0$ & $06.6$ & $07.7$ & \color{blue}{$05.9$} \\
Churn     & $15.0$ & $15.7$ & $19.9$ & \color{red}{$22.7$} & \color{blue}{$12.9$} & $16.2$ & $13.1$ & $13.0$ \\
COMPAS   & \color{red}{$18.3$} & \color{red}{$18.3$} & $17.7$ & $17.3$ & $17.9$ & $17.7$ & $17.2$ & \color{blue}{$17.0$} \\
Credit    & \color{red}{$26.9$} & \color{red}{$26.9$} & $12.4$ & $19.1$ & $19.4$ & \color{blue}{$12.2$} & $17.0$ & $15.8$ \\
Heart     & $28.7$ & $24.0$ & \color{red}{$32.6$} & \color{blue}{$15.4$} & $25.0$ & $30.8$ & $21.5$ & $21.8$ \\
MIMIC-II  & $20.5$ & $20.4$ & $19.4$ & \color{red}{$21.0$} & $19.6$ & $19.4$ & $18.8$ & \color{blue}{$18.6$} \\
MIMIC-III & $21.2$ & $20.7$ & $19.0$ & \color{red}{$21.6$} & $18.7$ & $18.7$ & $18.6$ & \color{blue}{$14.8$} \\
Pneumonia & \color{red}{$29.9$} & $29.7$ & $27.2$ & \color{blue}{$19.5$} & $25.3$ & $25.8$ & $25.8$ & $20.6$ \\
Support2  & $11.4$ & $12.4$ & \color{blue}{$10.3$} & $12.6$ & \color{red}{$13.0$} & $10.2$ & $11.4$ & $11.7$ \\ \bottomrule
Average   & \color{red}{$21.7$} & $20.4$ & $19.4$ & $19.3$ & $18.8$ & $17.9$ & $17.2$ & \color{blue}{$16.2$} \\
\end{tabular}

\label{table:l2ness}
\end{table}

In this section we propose a new metric to quantify feature sparsity, the notion that some GAM algorithms use fewer features than others to make predictions, which we have seen in Sec.~\ref{sec:fairness} to impact bias discovery.

\textbf{Feature density metric \ \ }
The idea is to quantify how fast the test \emph{error} of a trained model decays (i.e., how fast the model becomes more accurate) as we allow the model to have access to more features; a sparse model only requires a few important features to quickly reduce its test error, while a dense model needs more features to recover because it will have spread learned effects across more of the features. Using the GAM formulation as in Equation~\ref{eqn:gam}, we proceed as follows to compute this metric: first we keep only $f_0$ and measure the GAM's test set error as the initial error.
Then for each step out of $D$ steps, we greedily search over which feature $f_j(x_j)$, when added back to the model, reduces its validation error the most. We add that feature back and measure how the model's test error decreases. We save the test error as each subsequent feature is added, until $D$ features are added after $D$ steps, and plot test error against features. Finally we compute the feature density metric as the normalized area under this curve, treating the initial test error as $100$ and final error (with $D$ features) as $0$. We expect an extremely sparse model to have value close to $0$, and a dense model to have value close to $50$.

Table~\ref{table:l2ness} presents normalized feature density for different GAM algorithms on ten datasets. As expected, EBM consistently has higher density (less sparsity) than EBM-BF across datasets, as it uses more features by design. 
Similarly, XGB-L2 also has higher density than XGB, and LR is higher than LASSO.
This confirms that the feature density metric reflects what we want.
% Similar to EBM-BF and EBM, XGB has consistently lower density than XGB-L2 because adding the L2 bias to the otherwise greedy XGB algorithm causes it to spread learned effects more uniformly among the features.
FLAM has low feature density, which is unsurprising due to the $\ell_1$ penalty present in the method. Spline does not exhibit a clear pattern of feature density. For example, Spline has the smallest density on the Adult dataset but the largest density on the Breast dataset.

\textbf{Implications \ \ } The proposed feature density metric captures expected behavior.
We see lower feature density for methods that greedily select the next best feature (e.g. EBM-BF) or have penalties that regularize for sparsity (e.g. FLAM).
Methods that repeatedly cycle over all features (e.g. EBM) have higher feature density. 

% \setlength\tabcolsep{1pt} % default value: 6pt
% \begin{figure}[tbp]
%   \begin{small}
%   \begin{center}
% \begin{tabular}{ccc}
%   & Adult & Breast \\
%  \raisebox{3.5\normalbaselineskip}[0pt][0pt]{\rotatebox[origin=c]{90}{Relative Test Error}} &
%  \includegraphics[width=0.48\linewidth]{figures/l2ness/add_logloss_adult.pdf} &
%   \includegraphics[width=0.48\linewidth]{figures/l2ness/add_logloss_breast.pdf} \\
%   & Number of features added & Number of features added
%   \end{tabular} 
% \end{center}
% \end{small}

%   \caption{
%     Measuring feature sparsity. It shows how the test error (y-axis) decays as more features (x-axis) are given to the model in the Adult (left) and Breast (right) datasets. The steeper the curve is, the fewer features it depends on.
%   }
%   \label{fig:l2ness_fig}
% \end{figure}

\subsection{GAM data fidelity}
\label{sec:bias_var}

% \setlength\tabcolsep{1.5pt} % default value: 6pt
% \begin{figure}[tbp]
%   \begin{center}
% %   \resizebox{\textwidth}{!}{%
% \begin{tabular}{cc}
%   Bias + Noise
%   & Variance \\
%  \includegraphics[width=0.5\linewidth]{figures/bias_var_figs/mimicii_bias_0515.pdf}
%  & \includegraphics[width=0.5\linewidth]{figures/bias_var_figs/mimicii_variance_0515.pdf} \\
%   \end{tabular} 
% \end{center}
%   \vspace{-5pt}
%   \caption{
%      An example of bias-variance analysis of different GAMs in the MIMIC-II dataset.
%   }
%   \label{fig:bias_var_fig}
% \end{figure}

\begin{figure}
  \begin{center}
    \setlength\tabcolsep{2pt} % default value: 6pt
    \begin{tabular}{cc}
     \raisebox{7.\normalbaselineskip}[0pt][0pt]{\rotatebox[origin=c]{90}{Variance Rank}} &
     \includegraphics[width=0.65\linewidth]{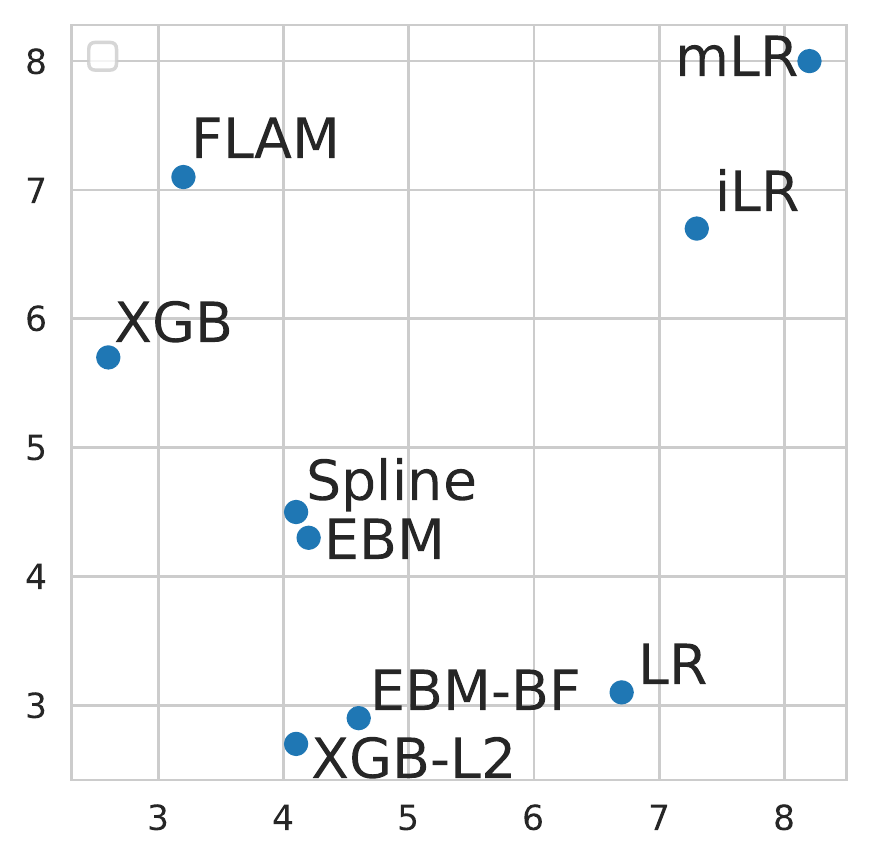} \\
      \ & Bias Rank
    \end{tabular} 
  \end{center}

  \caption{
     Bias rank (x-axis) vs. variance rank (y-axis) for each GAM across multiple datasets. Lower rank is better.
  }
  \label{fig:bias_var_summary_rank}
\end{figure}

In this section we propose a new metric to quantify how well a GAM is able to capture underlying data patterns, which we have seen in Sec. \ref{sec:anomaly} to impact data anomaly discovery. 

At first glance, one may think that test accuracy is a suitable metric for this purpose, since it captures how well a model generalizes to unseen data. However, we saw in Sec. \ref{sec:real_world} when comparing GAM algorithms of similar test accuracy how some were less able to represent certain data patterns. For example, smooth basis functions in Spline, while reducing variance and hopefully improving test set generalization, limited the model's ability to capture sharp jumps in the data. As noted by \citet{shmueli2010explain}, some highly accurate predictive models may actually be ``wrong'' in terms of capturing underlying data patterns. This notion is exactly statistical bias, which arises from model misspecification of the underlying data patterns~\citep{hastie2009elements}.  

% \textbf{Data fidelity metric \ \ } We use an approximation to the bias term in a bias-variance analysis of test set loss to measure data fidelity, which we estimate following \cite{munson2009feature, bauer1999empirical}: for each round, the dataset is split into 85-15\% train-test splits. The training data is subsampled to 50\% five different times, each time training a different model and collecting its test set predictions. Let $y_m$ be the average prediction for a particular test set point, averaged across the five models. We compute the loss between $y_m$ and $y$, the actual label for that point, over all test set points. This loss is our approximation to the bias term desired; it is an approximation because it includes intrinsic noise, which as noted in \citet{munson2009feature}, cannot be separated from bias for real datasets. Finally, the bias and variance estimates are averaged over eight rounds to increase robustness, and ranked compared to other GAM algorithms. For both bias and variance estimates, lower rank is better. Methods that have both low bias and variance (and thus low error) will fall closest to the bottom left corner of the graph.

\textbf{Data fidelity metric \ \ } We use an approximation to the bias term in a bias-variance analysis to measure data fidelity.
In bias-variance analysis~\citep{bauer1999empirical}, the loss of model is composed of noise $N(x)$, bias $B(x)$ and variance $V(x)$ terms:
$$
E_{D, t}[L(t, y)]=N(x)+B(x)+V(x) \text{\ \ \ \ where}
$$
$$
N(x) = E_{t}[L\left(t, y_{*}\right) ], B(x) = L\left(y_{*}, y_{m}\right), V(x)=E_{D}\left[L\left(y_{m}, y\right)\right]
$$
where $D$ is the training distribution, $t$ is the true label, $y_{*}$ is the optimal predictions, $y_m$ is the mean prediction of models across possible training datasets, and $y$ is the model.
Since we do not know the $y_{*}$, we instead measure the \emph{empirical bias} combining both noise and bias $N(x) + B(x) = E_{t}[L\left(t, y_{m}\right)]$ following~\citet{munson2009feature}.
We use the following sampling procedure: in each round, we split our dataset into 85-15\% train-test splits. 
We then randomly subsample the training data to $50\%$ and train models $5$ times, and we set the average of $5$ models as $y_{m}$ to calculate empirical bias and variance once.
Finally, the bias and variance estimates are averaged over eight rounds, and ranked compared to other GAM algorithms on each dataset.
We take the average ranks across the ten datasets (lower rank is better).

Fig.~\ref{fig:bias_var_summary_rank} plots average variance rank vs. average bias rank for different GAM algorithms. 
Considering GAM algorithms closest to the bottom left corner (i.e. (0, 0) point), which are also the most accurate GAMs (see Table \ref{table:real_data_test_AUC}), XGB has the highest data fidelity (lowest bias rank) but has rather high variance.
%but high variance rank (high variance) and thus should be the most trustworthy, though its high variance might make it more difficult to distinguish real from spurious patterns. % , while having higher variance could confuse the patterns
FLAM has the next highest data fidelity, but has even higher variance, hence it is dominated by XGB that has both higher data fidelity and lower variance.
After FLAM, XGB-L2, Spline, and EBM have the next highest data fidelity, and promisingly, with significantly lower variance than FLAM or XGB.

%Thus shape plots from EBM-BF should be less trusted especially given that EBM-BF does not have higher accuracy than the other methods. 

\textbf{Implications \ \ } 
We use statistical bias as a proxy to measure data fidelity with real data.
By decomposing error into bias and variance components, we see that equally accurate GAM algorithms achieve the same accuracy in different ways. Certain GAM algorithms (e.g. XGB) have lower bias which indicates better fidelity, while other GAMs (e.g. XGB-L2) have lower variance at the expense of higher bias. 
% By decomposing error into bias and variance components, we see that equally accurate GAM algorithms achieve the same accuracy in different ways, with certain GAM algorithms (Spline, EBM) roughly trading off bias and variance equally while other GAM algorithms focus on one component at the expense of the other. 

%We use statistical bias as a proxy to measure data fidelity with real data.
%Across ten datasets we experiment with, patterns learned by XGB and FLAM are probably the most trustworthy, though their relatively high variance might create some difficulties for interpretations.
%Spline and EBM have more bias but less variance, and XGB-L2 exhibits very low variance for little increase in bias. 
%EBM-BF is dominated by XGB-L2 and has higher bias, and thus is the least trustworthy of the GAM algorithms.  
% In summary, XGB, Spline and EBM appear to be  best methods overall.
% In summary, XGB and FLAM are the most trustworthy with the smallest bias. Spline and EBM are similar in the middle ground with EBM-BF as the most untrustworthy. 

\subsection{GAM data fidelity and generator bias}
\label{sec:ss_graph_fidelity}

\setlength\tabcolsep{0pt} % default value: 6pt
\begin{figure*}[tbp]
  \begin{center}

\begin{tabular}{ccccc}
   & \multicolumn{2}{c}{Systolic BP} & \multicolumn{2}{c}{AIDS} \\
     \cmidrule(lr){2-3}\cmidrule(lr){4-5}
   & (a) Spline generator
   & (b) FLAM generator
   & (c) Spline generator
   & (d) FLAM generator \\
 \raisebox{4.\normalbaselineskip}[0pt][0pt]{\rotatebox[origin=c]{90}{\small Log odds}}
 & \includegraphics[width=0.25\linewidth]{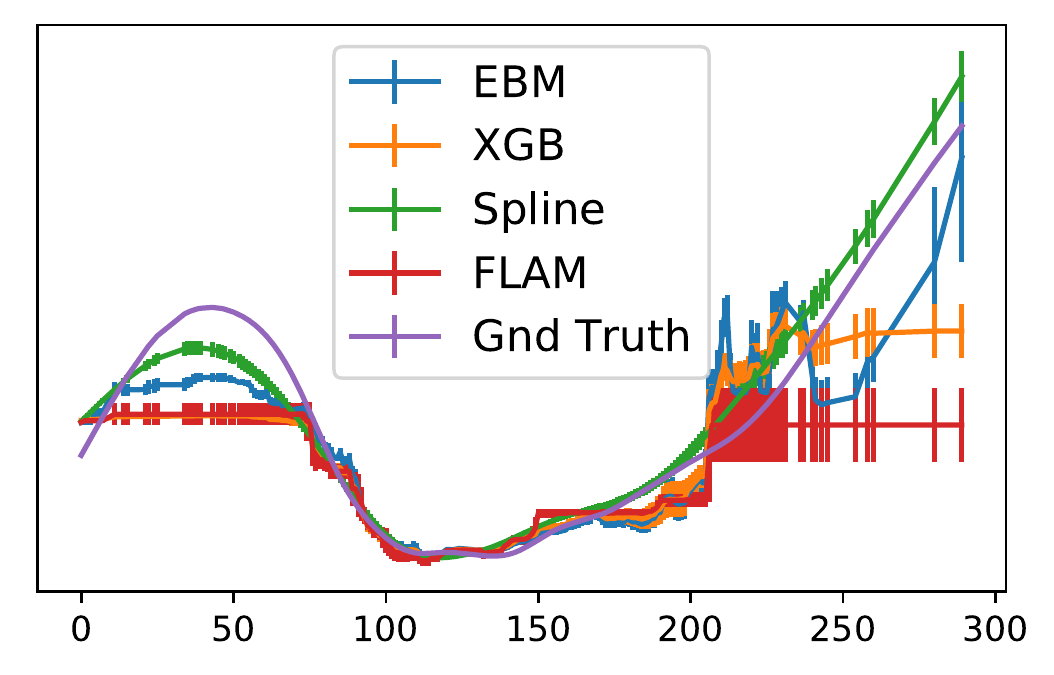}
 & \includegraphics[width=0.25\linewidth]{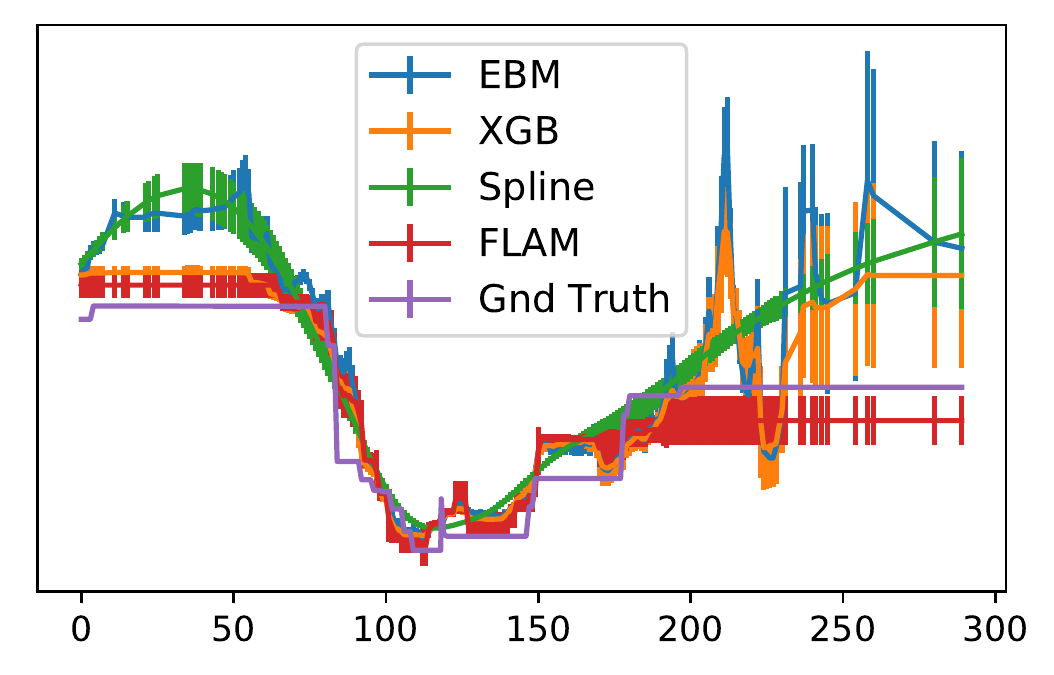}
% \\ 
%  \raisebox{3. \normalbaselineskip}[0pt][0pt]{\rotatebox[origin=c]{90}{FLAM generator}}
 & \includegraphics[width=0.25\linewidth]{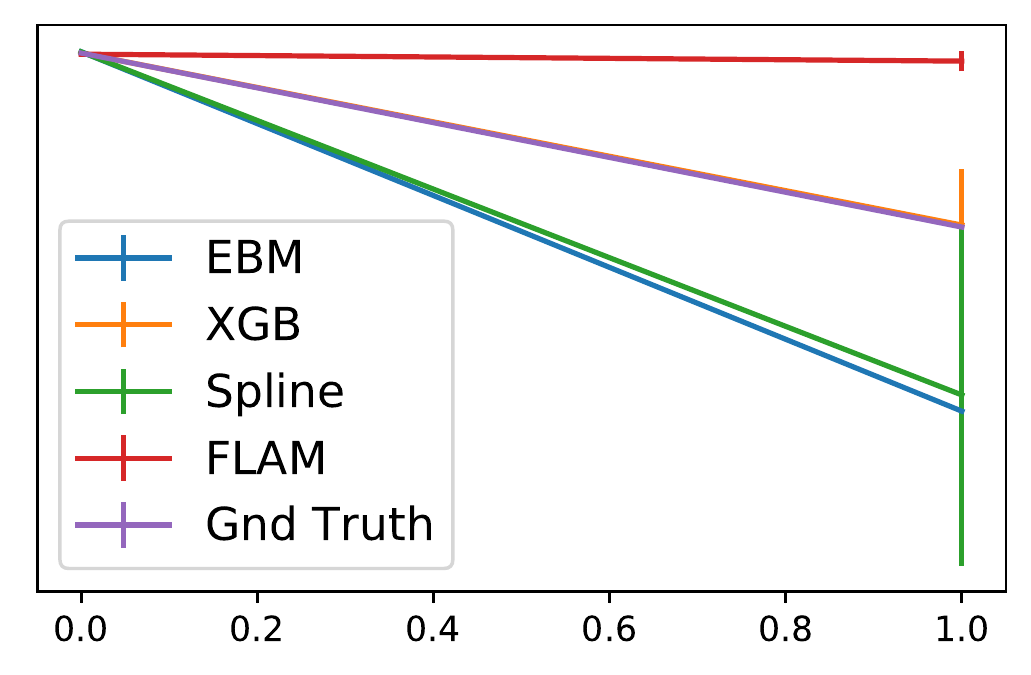}
 & \includegraphics[width=0.25\linewidth]{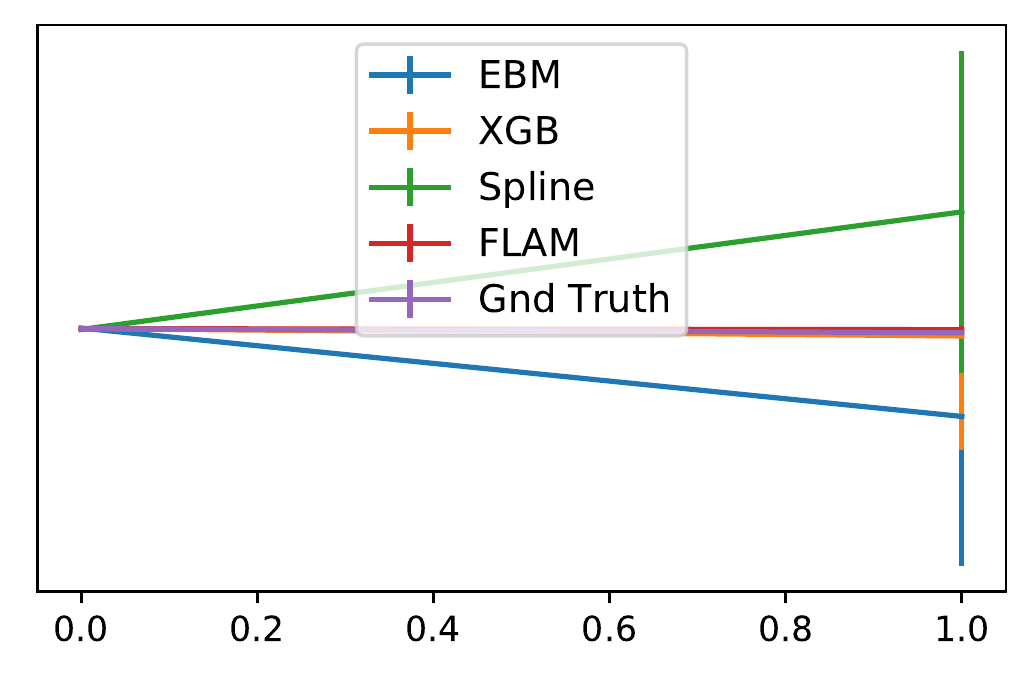}
 \\
  \end{tabular}
  
\end{center}

  \caption{
     Shape plots for Systolic BP and AIDS features in semi-synthetic MIMIC-II, generated using different generators (Spline and FLAM). 
  }
  \label{fig:ss_different_generators}
\end{figure*}

We have thus far studied the data fidelity properties of different GAM algorithms on several real datasets. However, it may be that the inductive bias of a certain GAM algorithm happened to agree with the (unknown) data pattern in a particular real dataset. In this section, we experiment with semi-synthetic datasets created using known data generators. To preserve the character of real datasets as much as possible, we keep the features $X$ but change the label $y$ by training multiple ground truth GAM models (EBM, XGB, Spline, FLAM and LR) on features $X$ and then \emph{re-generating} the label $y$ as each model's predictions.
Since these GAM models (except LR) are among the most accurate models on most datasets (Table~\ref{table:real_data_test_AUC}), the generated labels capture the real-world distribution as close as possible.
As these GAM algorithms are very different from each other, this should provide a diversity of ground truth data patterns.

Fig.~\ref{fig:ss_different_generators}(a)-(d) shows different GAMs alongside ground truth patterns from two very different generators, Spline and FLAM, on MIMIC-II for one continuous feature (Systolic BP) and one boolean feature (AIDS). Purple represents ground truth, i.e. Spline generator for Fig.~\ref{fig:ss_different_generators}(a) and (c), and FLAM generator for ~\ref{fig:ss_different_generators}(b) and (d). We see an obvious \textit{generator bias}: a GAM algorithm fits the ground truth better when ground truth is generated using the same algorithm. For example, on Systolic BP, the Spline GAM fits well the data generated by its own generator (Fig.~\ref{fig:ss_different_generators}(a)), while doing poorly for data generated by the FLAM generator (Fig.~\ref{fig:ss_different_generators}(b)), and vice versa for FLAM. However, tree-based methods (EBM, XGB) on Systolic BP with the Spline generator (Fig.~\ref{fig:ss_different_generators}a) still learn abrupt jumps at $225$ even when the underlying ground truth is smooth; similarly there is also a drop at $175$.
This illustrates that it is possible for model inductive bias to dominate irrespective of the true data generator.

To mitigate the aforementioned generator bias, we perform a worst-case analysis: what is the worst performance each GAM algorithm would get across all of the different data generators?
Since we do not know the underlying generators on real datasets -- they could be jumpy, smooth, or even linear -- this analysis is more realistic and robust to all these cases.
%As in the real world, we do not know properties of the underlying generator in advance, as they can be jumpy, smooth or even linear, and this analysis helps us understand how poorly each GAM algorithm can do.

\textbf{Worst-case data fidelity metric \ \ } 
% %Specifically, we simulate datasets with known ground truth shape plots and see how which GAM algorithms can best recover the underlying plots across a variety of data generating processes. 
To measure how well a GAM can recover the ground truth generators, we calculate the mean absolute difference of each shape plot between the ground truth GAM and the GAM model. Specifically, using the GAM formulation as in Equation \ref{eqn:gam} where $f_j$ is the shape function for feature $j$, and taking $g_j$ to be the shape function for the ground truth GAM, we calculate the absolute difference $\sum_{j=1}^D |f_j(x_j) - g_j(x_j)|$ across the whole dataset.
% , and sum the error weighted by number of examples.
% This measures how good a method recovers the generator.
%Note that this is different from test set error, as one method could have very high data infidelity for some or parts of the shape function, but with correlated features might still have high test set accuracy.
To compare between datasets, we linearly scale the absolute difference between 0 and 100 for a particular semi-synthetic dataset, with the worst GAM algorithm having value 0 and best GAM algorithm having value 100. We then take the worst score over the five different data generators that yielded five semi-synthetic datasets from each real dataset.

\setlength\tabcolsep{1pt} % default value: 6pt
\begin{table}[tbp]
\centering
\caption{Worst-case data fidelity ($\%$) taking into account different data generators. Each row aggregates the results over five different generators (EBM, XGB, FLAM, Spline and LR). Higher numbers are better. Best number in each row is in \textbf{bold}.}
\begin{tabular}{ccccccccc}
& FLAM        & XGB         & EBM    & Spline & EBM-BF & LR          & iLR    & mLR   \\ \toprule
Breast    & $22.9$      & $30.3$      & $0$  & $13.3$ & $21.2$ & $\bm{42.5}$ & $0$  & $0$ \\
Churn     & $\bm{20.3}$ & $10.5$      & $13.5$ & $0$  & $1.3$  & $0$       & $16.0$ & $0$ \\
Heart     & $\bm{86.9}$ & $68.2$      & $68.7$ & $69.7$ & $24.8$ & $52.4$      & $64.6$ & $0$ \\
MIMIC-II  & $62.9$      & $\bm{73.9}$ & $61.2$ & $72.7$ & $52.6$ & $0$       & $6.6$  & $0$ \\
MIMIC-III & $65.2$      & $\bm{70.1}$ & $45.3$ & $51.2$ & $37.0$ & $27.0$      & $0$  & $0$ \\
Pneumonia & $\bm{64.4}$ & $60.2$      & $40.0$ & $3.6$  & $0$  & $0$       & $26.8$ & $6.4$ \\ \bottomrule
Average   & $\bm{53.8}$ & $52.2$      & $38.1$ & $35.1$ & $22.8$ & $20.3$      & $19.0$ & $1.1$
%  & EBM & EBM-BF & XGB & FLAM & Spline & LR & iLR & mLR \\ \toprule
% Breast & $0$ & $21.2$ & $30.3$ & $22.9$ & $13.3$ & $\bm{42.5}$ & $0$ & $0$ \\
% Churn & $13.5$ & $1.3$ & $10.5$ & $\bm{20.3}$ & $0$ & $0$ & $16.0$ & $0$ \\
% Heart & $68.7$ & $24.8$ & $68.2$ & $\bm{86.9}$ & $69.7$ & $52.4$ & $64.6$ & $0$ \\
% MIMIC-II & $61.2$ & $52.6$ & $\bm{73.9}$ & $62.9$ & $72.7$ & $0$ & $6.6$ & $0$ \\
% MIMIC-III & $45.3$ & $37.0$ & $\bm{70.1}$ & $65.2$ & $51.2$ & $27.0$ & $0$ & $0$ \\
% Pneumonia & $40.0$ & $0$ & $60.2$ & $\bm{64.4}$ & $3.6$ & $0$ & $26.8$ & $6.4$ \\ \bottomrule
% Average & $38.1$ & $22.8$ & $52.2$ & $\bm{53.8}$ & $35.1$ & $20.3$ & $19.0$ & $1.1$
\end{tabular}

\label{table:graph_fidelity}
\end{table}

Table~\ref{table:graph_fidelity} provides the worst-case data fidelity for eight GAM algorithms on six real datasets, where each dataset (row) encapsulates five semi-synthetic datasets from different data generators. FLAM and XGB performed the best, then EBM and Spline. 

\textbf{Implications \ \ } FLAM and XGB exhibit the best worst-case data fidelity. Spline and EBM are similar, and  EBM-BF is the worst. Taking into account different data generators, our results are not substantively different from the results derived from the bias-variance analysis on real data in Sec.~\ref{sec:bias_var}.

%FLAM and XGB exhibit the best worst-case data fidelity.
%Spline and EBM are similar, and  EBM-BF is the worst.
%Note that this ranking is very similar to the results we derived from the bias-variance analysis in real data (Sec.~\ref{sec:bias_var}).

% This ranking match exactly with the bias rank in the bias-variance analysis, implying that the higher explanation fidelity implies lower bias of the model.
% This ranking match exactly with the bias rank in the bias-variance analysis, that the higher explanation fidelity implies with low bias of the model.
% This is exactly aligned with the bias-variance analysis (Sec. \ref{sec:bias_var}) that XGB and FLAM also having smallest bias that resembles the ground truth more faithfully.

\subsection{GAM accuracy vs. data fidelity}
\label{sec:ss_correlation_btw_explanation_and_generalization}

% \begin{figure*}[tbp]
%   \begin{center}
%     \begin{tabular}{cccccc}
%       \ & (a) EBM & (b) XGB & (c) FLAM & (d) Spline & (e) LR \\
%      \raisebox{3.3\normalbaselineskip}[0pt][0pt]{\rotatebox[origin=c]{90}{Data fidelity rank}} &
%      \includegraphics[width=0.19\linewidth]{figures/ss/SS_corr_EBM.pdf} &
%      \includegraphics[width=0.19\linewidth]{figures/ss/SS_corr_XGB.pdf} &
%      \includegraphics[width=0.19\linewidth]{figures/ss/SS_corr_FLAM.pdf} &
%      \includegraphics[width=0.19\linewidth]{figures/ss/SS_corr_Spline.pdf} &
%       \includegraphics[width=0.19\linewidth]{figures/ss/SS_corr_LR.pdf} \\
%      \ & Test AUC rank & Test AUC rank & Test AUC rank & Test AUC rank & Test AUC rank
%     \end{tabular} 
%   \end{center}

%   \caption{
%       Scatter plot between the ranking of test AUC (x-axis) vs the ranking of data fidelity (y-axis) for (a) EBM, (b) XGB, (c) FLAM, (d) Spline and (e) LR. Color specifies different generators, and density indicates the number of points. Lower rank is better. The more points in the upper left corner (low test AUC rank but high data fidelity rank), the more unreliable of using high test accuracy to indicate good data fidelity.
%   }
%   \label{fig:ss_test_auc_graph_mae_rank}
% %   \vspace{-4pt}
% \end{figure*}

A GAM model that has high accuracy but low data fidelity may mislead users who tend to judge models solely based on accuracy. We quantify which GAM algorithm is more likely to mislead users this way, by comparing the difference between test AUC rank and data fidelity rank. For each dataset, we compute these two ranks as in Sec. \ref{sec:dataset} and Sec. \ref{sec:bias_var}, with lower rank being better. Then we take the rank of fidelity minus the rank of test AUC. If the result is negative, we clip it at $0$. We call this the ``positive difference" between the two ranks. Finally, we average this over all thirty semi-synthetic datasets. We expect a misleading model to have a lower test AUC rank and higher data fidelity rank.

% Since we don't care when the model has low test AUC but high fidelity, we only take the positive difference between the test AUC ranking and fidelity (i.e. treating negative value as $0$).
% Then we only take the positive difference between the rankings (i.e. treating negative value as $0$), since we don't care the other side when the model has low test AUC but high graph fidelity.
% Then we take the largest and average difference among all $30$ simulated datasets.
% To compare all GAMs, for each semi-synthetic data out of $30$ we take only the positive difference between the relative ranking of test AUC and data fidelity (Sec.~\ref{sec:ss_graph_fidelity}).

From Table~\ref{table:corr_test_acc_graph}, Spline has the largest difference in rank over multiple datasets with different data generators. This rank difference is largest when the data generators are jumpy, which creates challenges for Spline which uses smooth basis functions.

% Examining which dataset causes this behavior, we find Spline 
% Specifically, we fine Spline has the largest rank discrepancy when generators are jumpy such as FLAM, XGB and EBM.
% while others have less and similar misleading behaviors.
% To investigate how correlated test accuracy is with data fidelity for each GAM method.
% , we scatter plot the ranking of the test AUC versus the ranking of data fidelity on all $30$ semi-synthetic data in Fig.~\ref{fig:ss_test_auc_graph_mae_rank}.
% In Fig.~\ref{fig:ss_test_auc_graph_mae_rank},  XGB (b) and LR (e) are two best methods with fewer points closer to the upper left corner, with EBM in the third place. FLAM and Spline are the two worst methods with points closer to the corner.
% Intuitively, methods tend to have the largest rank discrepancy when the generator has very different inductive bias.
% Indeed, FLAM (Figure \ref{fig:ss_test_auc_graph_mae_rank}(c)) does the worst with the LR generator (orange) where it gets rank $1$ in test AUC yet has rank $6$ in data fidelity.
% Similarly, Spline (Figure \ref{fig:ss_test_auc_graph_mae_rank}(d)) has the largest rank discrepancy when generators are jumpy such as FLAM, XGB and EBM.

\textbf{Implications \ \ }
For Spline, using high test accuracy alone to select a model may be misleading, especially when the underlying data pattern may be jumpy. Other methods are more stable.
% Tree-based method XGB and EBM and linear method LR are more stable and have better correlation between test accuracy and data fidelity.

\setlength\tabcolsep{3pt} % default value: 6pt
\begin{table}[tbp]
    \centering
    \caption{Difference between test AUC rank and data fidelity rank on thirty semi-synthetic datasets. The larger this difference, the less reliable it is to use high accuracy to infer good data fidelity. Best number is \textbf{bold}.}
    \label{table:corr_test_acc_graph}
\begin{tabular}{c|cccccc}
 & EBM & FLAM & XGB & LR & EBM-BF & Spline   \\ \toprule
% Max Diff in Rank     & $5$  & $5$  & $5$    & $4$    & $6$      & $\bm{3}$    \\
Avg Pos Diff in Rank     & $\bm{0.47}$ & $0.50$ &  $0.62$ & $0.63$ & $0.87$  & $1.22$ \\
\end{tabular}
\end{table}

% As a result, for methods such as Spline or FLAM having large discrepency, using low generalization error to indicate good explanation could be misleading, especially when the model inductive bias disagrees with the data.

% , and caculate the $R^2$ to measure the correlation.

% The top $2$ correlated method is LR ($0.61$) and XGB ($0.51$), with EBM coming in the third place ($0.23$).
% The poorest correlated methods are Spline ($0.16$) and FLAM ($0.08$).
% As a result, for poorly correlated methods we might be cautious of using the low generalization error to indicate good graph fidelity.
% TODO: Any deduction of why it happens?

% \subsection{Runtime analysis}

\section{Discussion}

GAMs are widely used to discover patterns in data in a variety of fields including business~\citep{sapra2013generalized}, healthcare~\citep{hunter2016have}, ecology~\citep{pedersen2019hierarchical}, horticulture~\citep{saw2017applying}, air pollution~\citep{ravindra2019generalized}, nutrition~\citep{rostami2020optimal} and  COVID-19~\citep{Izadi2020generalized}. But most of these research only experimented with a specific GAM algorithm  (typically Spline) without any comparison to other GAM algorithms. In this work, we have shown that the patterns learned by GAMs are highly impacted by their own inductive biases. 
If the papers that used GAMs to discover patterns had used different GAM algorithms, would they have drawn different conclusions?
How many of the findings are due to true patterns in the data and not due to the inductive bias of the particular GAM algorithm chosen?

%%% Important paragraphs: what is its implication for fairness people? I probably should add back the conclusion that to avoid the model inductive bias that biases the explanations, we should run multiple GAMs and believe the common patterns, and believe tree-based GAMs when conflicts occur.

While we aimed to provide a useful and fair experimental study, there are limitations to the conclusions that can be drawn from our work due to design choices we made. In terms of data sets, we considered common Kaggle datasets across several domains that are relatively large but still have a manageable amount of features. We do not explore small datasets used in the Spline literature, where a smoothing prior might help compensate for a lack of sample size.
In terms of models, we only focused on a few of the most representative GAM algrithms and make additional modifications to these methods to study different characteristics of GAMs (e.g. feature sparsity and data fidelity). We leave more theoretical comparisons to future work.
% We also do not compare pairwise interaction effects in GA$^2$M which are useful in dataset bias discovery~\citep{lou2013accurate, caruana2015intelligible}.
% There are also many small details to consider (categorical feature encoding, hyperparameter selection).
% We also .
%We take our best effort to compare various packages for the same method (especially for Spline) and make our best judgements for small details such as categorical feature encoding and hyperparameter selections to make fair comparisons.
% Implementing the different GAM methods has proven to be a difficult endeavour. 
% Extracting underlying graphs from different package in both Python and R languange is not an easy route.

%% TODO: pargraph linked back to fairness that drives home!
% Real examples that GAM-based justice model.
% Make the reader feel like this is actually very important
% This is not just academic exploration. In real world, people actually use this.
% Discussion about the bias (various definition of bias). We use the loss of different subpopulation is the various bias.

% \note{May remove it to save space}
% There are many ways to measure bias and fairness.  In this paper we used one metric --- loss on minority vs majority subpopulations --- to demonstrate how GAMs trained with different algorithms can learn different patterns from the data that might selectively disadvantage certain groups. The results suggest that the different inductive biases exhibited by different GAM methods can lead to different biases with respect to under represented classes in the data.

\setlength\tabcolsep{3pt} % default value: 6pt
\begin{table}[tbp]
    \centering
    \caption{Summary of key findings, ranking the different GAM algorithms across six properties studied in this paper. Best number in each row is in \textbf{bold} or \textcolor{red}{red}.
    }
    \label{table:summary_findings}
\begin{tabular}{c|ccccc}
 & EBM & XGB & FLAM & Spline & LR \\ \toprule

\makecell{Test-set\\accuracy} & $\bm{1.5}$ & $\bm{1.5}$ & 4 & 3 & 5 \\ \hline 

\makecell{Feature\\density} & $\bm{1}$ & 4 & 5 & 2.5 & 2.5 \\  \hline
\makecell{Low bias\\(bias/variance)} & 3.5 & $\bm{1}$ & 2 & 3.5 & 5 \\ \hline
\makecell{Worst-case\\fidelity} & 3.5 & $\bm{1.5}$ & $\bm{1.5}$ & 3.5 & 5 \\ \hline
\makecell{High accuracy\\implies good data fidelity} & $\bm{1.5}$ & 3.5 & $\bm{1.5}$ & 5 & 3.5 \\ \hline
\makecell{Anomaly\\detection} & $\bm{1.5}$ & $\bm{1.5}$ & 3 & 4.5 & 4.5 \\ \bottomrule
Sum of Ranks & \color{red}{$\bm{12.5}$} & \color{red}{$\bm{13}$} & 17 & 22 & 25.5 \\
% Average & \bm{2} & 2.2 & 3.2 & 3.9 & 4 & \\
\end{tabular}
\end{table}

\section{Conclusion}
\label{sec:conclusion}
The key findings are summarized in Table~\ref{table:summary_findings}, where we have synthesized our findings across six different properties studied in this paper and ranked each GAM algorithm for each property (ties count for half a rank). Although a number of GAM algorithms yield similar accuracy, tree-based methods like EBM and XGB are superior when considering issues such as bias and data anomaly discovery, sparsity, fidelity, and accuracy.  
Tree-based methods such as XGB and EBM have higher feature density than FLAM or Spline. They also have less bias on real data, and recover data patterns with better fidelity on semi-synthetic data.
We also find Spline could have high accuracy yet at the same time low data fidelity, which might mislead users who perform model selection based on test accuracy alone. 
% We also find tree-based algorithms have better correlation between test accuracy and data fidelity, making them more reliable when performing model selection based on test accuracy.
Qualitatively, Spline and FLAM are not good at detecting local anomalies such as mean imputation or treatment effects, both of which are easily detected by the tree-based methods.
Spline also extrapolates over-confidently in low-sample regions (Fig.~\ref{fig:poster_child}(b), and see other examples in Appendix~\ref{sec:appx_additional_shape_graphs}).

% Among the tree-based methods, EBM's higher density makes it easier to discover bias hidden in the data. Yet XGB is, on average, more faithful to data generating process in both real and semi-synthetic datasets.
% Overall, the safest bet is probably to run different GAMs, both smooth and non-smooth, carefully examine all their explanations, and trust XGB and EBM more when conflicts occur.

Future development of better GAM algorithms should focus on the following:
(1) GAMs that can better capture rapid non-linear change, (2) GAMs with high feature density to improve fairness and prevent bias masking, (3) GAMs having higher data fidelity on both real and simulated data.
We believe our work is an important step towards making GAMs more trustworthy, and our evaluation framework will promote the development of better GAMs in the future.

\begin{acks}
This work was created during an internship at Microsoft Research. Resources used in preparing this research were provided, in part, by the Province of Ontario, the Government of Canada through CIFAR, and companies sponsoring the Vector Institute (\url{www.vectorinstitute.ai/\#partners}).
\end{acks}

\begin{small}
\bibliographystyle{ACM-Reference-Format}
\bibliography{facct}


\begin{thebibliography}{45}

%%% the Web address.

\ifx \showCODEN    \undefined \def \showCODEN     #1{\unskip}     \fi
\ifx \showDOI      \undefined \def \showDOI       #1{#1}\fi
\ifx \showISBNx    \undefined \def \showISBNx     #1{\unskip}     \fi
\ifx \showISBNxiii \undefined \def \showISBNxiii  #1{\unskip}     \fi
\ifx \showISSN     \undefined \def \showISSN      #1{\unskip}     \fi
\ifx \showLCCN     \undefined \def \showLCCN      #1{\unskip}     \fi
\ifx \shownote     \undefined \def \shownote      #1{#1}          \fi
\ifx \showarticletitle \undefined \def \showarticletitle #1{#1}   \fi
\ifx \showURL      \undefined \def \showURL       {\relax}        \fi
% invisible to TeX
\providecommand\bibfield[2]{#2}
\providecommand\bibinfo[2]{#2}
\providecommand\natexlab[1]{#1}
\providecommand\showeprint[2][]{arXiv:#2}

\bibitem[\protect\citeauthoryear{Angwin, Larson, Mattu, and Kirchner}{Angwin
  et~al\mbox{.}}{2019}]%
        {propublica}
\bibfield{author}{\bibinfo{person}{Julia Angwin}, \bibinfo{person}{Jeff
  Larson}, \bibinfo{person}{Surya Mattu}, {and} \bibinfo{person}{Lauren
  Kirchner}.} \bibinfo{year}{2019}\natexlab{}.
\newblock \showarticletitle{Machine Bias: There’s software used across the
  country to predict future criminals. And it’s biased against blacks.}
\newblock  (\bibinfo{year}{2019}).
\newblock
\urldef\tempurl%
\url{https://www.propublica.org/article/machine-biasrisk-assessments-in-criminal-sentencing}
\showURL{%
\tempurl}


\bibitem[\protect\citeauthoryear{Bauer and Kohavi}{Bauer and Kohavi}{1999}]%
        {bauer1999empirical}
\bibfield{author}{\bibinfo{person}{Eric Bauer} {and} \bibinfo{person}{Ron
  Kohavi}.} \bibinfo{year}{1999}\natexlab{}.
\newblock \showarticletitle{An empirical comparison of voting classification
  algorithms: Bagging, boosting, and variants}.
\newblock \bibinfo{journal}{\emph{Machine learning}} \bibinfo{volume}{36},
  \bibinfo{number}{1} (\bibinfo{year}{1999}).
\newblock


\bibitem[\protect\citeauthoryear{Binder and Tutz}{Binder and Tutz}{2008}]%
        {binder2008comparison}
\bibfield{author}{\bibinfo{person}{Harald Binder} {and}
  \bibinfo{person}{Gerhard Tutz}.} \bibinfo{year}{2008}\natexlab{}.
\newblock \showarticletitle{A comparison of methods for the fitting of
  generalized additive models}.
\newblock \bibinfo{journal}{\emph{Statistics and Computing}}
  \bibinfo{volume}{18}, \bibinfo{number}{1} (\bibinfo{year}{2008}).
\newblock


\bibitem[\protect\citeauthoryear{Caruana, Lou, Gehrke, Koch, Sturm, and
  Elhadad}{Caruana et~al\mbox{.}}{2015}]%
        {caruana2015intelligible}
\bibfield{author}{\bibinfo{person}{Rich Caruana}, \bibinfo{person}{Yin Lou},
  \bibinfo{person}{Johannes Gehrke}, \bibinfo{person}{Paul Koch},
  \bibinfo{person}{Marc Sturm}, {and} \bibinfo{person}{Noemie Elhadad}.}
  \bibinfo{year}{2015}\natexlab{}.
\newblock \showarticletitle{Intelligible models for healthcare: Predicting
  pneumonia risk and hospital 30-day readmission}. In
  \bibinfo{booktitle}{\emph{KDD}}.
\newblock


\bibitem[\protect\citeauthoryear{Chen and Guestrin}{Chen and Guestrin}{2016}]%
        {Chen:2016:XST:2939672.2939785}
\bibfield{author}{\bibinfo{person}{Tianqi Chen} {and} \bibinfo{person}{Carlos
  Guestrin}.} \bibinfo{year}{2016}\natexlab{}.
\newblock \showarticletitle{{XGBoost}: A Scalable Tree Boosting System}. In
  \bibinfo{booktitle}{\emph{KDD}}.
\newblock


\bibitem[\protect\citeauthoryear{Chouldechova}{Chouldechova}{2017}]%
        {chouldechova2017fair}
\bibfield{author}{\bibinfo{person}{Alexandra Chouldechova}.}
  \bibinfo{year}{2017}\natexlab{}.
\newblock \showarticletitle{Fair prediction with disparate impact: A study of
  bias in recidivism prediction instruments}.
\newblock \bibinfo{journal}{\emph{Big Data}} \bibinfo{volume}{5},
  \bibinfo{number}{2} (\bibinfo{year}{2017}).
\newblock


\bibitem[\protect\citeauthoryear{Doshi-Velez and Kim}{Doshi-Velez and
  Kim}{2017}]%
        {doshi2017towards}
\bibfield{author}{\bibinfo{person}{Finale Doshi-Velez} {and}
  \bibinfo{person}{Been Kim}.} \bibinfo{year}{2017}\natexlab{}.
\newblock \showarticletitle{Towards A Rigorous Science of Interpretable Machine
  Learning}.
\newblock \bibinfo{journal}{\emph{Springer Series on Challenges in Machine
  Learning: "Explainable and Interpretable Models in Computer Vision and
  Machine Learning"}} (\bibinfo{year}{2017}).
\newblock


\bibitem[\protect\citeauthoryear{Dua and Graff}{Dua and Graff}{2017}]%
        {UCI}
\bibfield{author}{\bibinfo{person}{Dheeru Dua} {and} \bibinfo{person}{Casey
  Graff}.} \bibinfo{year}{2017}\natexlab{}.
\newblock \bibinfo{title}{{UCI} Machine Learning Repository}.
\newblock
\newblock
\urldef\tempurl%
\url{http://archive.ics.uci.edu/ml}
\showURL{%
\tempurl}


\bibitem[\protect\citeauthoryear{Dwyer and Holte}{Dwyer and Holte}{2007}]%
        {dwyer2007instability}
\bibfield{author}{\bibinfo{person}{Kenneth Dwyer} {and} \bibinfo{person}{Robert
  Holte}.} \bibinfo{year}{2007}\natexlab{}.
\newblock \showarticletitle{Decision Tree Instability and Active Learning}. In
  \bibinfo{booktitle}{\emph{ECML}}.
\newblock


\bibitem[\protect\citeauthoryear{Hastie and Tibshirani}{Hastie and
  Tibshirani}{1990}]%
        {hastie1990generalized}
\bibfield{author}{\bibinfo{person}{Trevor Hastie} {and} \bibinfo{person}{Rob
  Tibshirani}.} \bibinfo{year}{1990}\natexlab{}.
\newblock \bibinfo{booktitle}{\emph{Generalized Additive Models}}.
\newblock \bibinfo{publisher}{Chapman and Hall/CRC}.
\newblock


\bibitem[\protect\citeauthoryear{Hastie and Tibshirani}{Hastie and
  Tibshirani}{1995}]%
        {hastie1995generalized}
\bibfield{author}{\bibinfo{person}{Trevor Hastie} {and} \bibinfo{person}{Robert
  Tibshirani}.} \bibinfo{year}{1995}\natexlab{}.
\newblock \showarticletitle{Generalized additive models for medical research}.
\newblock \bibinfo{journal}{\emph{Statistical Methods in Medical Research}}
  \bibinfo{volume}{4}, \bibinfo{number}{3} (\bibinfo{year}{1995}).
\newblock


\bibitem[\protect\citeauthoryear{Hastie, Tibshirani, and Friedman}{Hastie
  et~al\mbox{.}}{2009}]%
        {hastie2009elements}
\bibfield{author}{\bibinfo{person}{Trevor Hastie}, \bibinfo{person}{Robert
  Tibshirani}, {and} \bibinfo{person}{Jerome Friedman}.}
  \bibinfo{year}{2009}\natexlab{}.
\newblock \bibinfo{booktitle}{\emph{The elements of statistical learning: data
  mining, inference, and prediction}}.
\newblock \bibinfo{publisher}{Springer Science \& Business Media}.
\newblock


\bibitem[\protect\citeauthoryear{Hegselmann, Volkert, Ohlenburg, Gottschalk,
  Dugas, and Ertmer}{Hegselmann et~al\mbox{.}}{2020}]%
        {doctor2020interpretable}
\bibfield{author}{\bibinfo{person}{Stefan Hegselmann}, \bibinfo{person}{Thomas
  Volkert}, \bibinfo{person}{Hendrik Ohlenburg}, \bibinfo{person}{Antje
  Gottschalk}, \bibinfo{person}{Martin Dugas}, {and} \bibinfo{person}{Christian
  Ertmer}.} \bibinfo{year}{2020}\natexlab{}.
\newblock \showarticletitle{An Evaluation of the Doctor-Interpretability of
  Generalized Additive Models with Interactions}. In
  \bibinfo{booktitle}{\emph{Machine Learning for Healthcare Conference}}.
\newblock


\bibitem[\protect\citeauthoryear{Hooker and Mentch}{Hooker and Mentch}{2019}]%
        {hooker2019permuting}
\bibfield{author}{\bibinfo{person}{Giles Hooker} {and} \bibinfo{person}{Lucas
  Mentch}.} \bibinfo{year}{2019}\natexlab{}.
\newblock \showarticletitle{Please Stop Permuting Features: An Explanation and
  Alternatives}.
\newblock \bibinfo{journal}{\emph{arXiv preprint arXiv:1905.03151}}
  (\bibinfo{year}{2019}).
\newblock


\bibitem[\protect\citeauthoryear{Hunter and Pr{\"u}ss-Ust{\"u}n}{Hunter and
  Pr{\"u}ss-Ust{\"u}n}{2016}]%
        {hunter2016have}
\bibfield{author}{\bibinfo{person}{Paul~R Hunter} {and}
  \bibinfo{person}{Annette Pr{\"u}ss-Ust{\"u}n}.}
  \bibinfo{year}{2016}\natexlab{}.
\newblock \showarticletitle{Have we substantially underestimated the impact of
  improved sanitation coverage on child health? A generalized additive model
  panel analysis of global data on child mortality and malnutrition}.
\newblock \bibinfo{journal}{\emph{PloS One}} \bibinfo{volume}{11},
  \bibinfo{number}{10} (\bibinfo{year}{2016}).
\newblock


\bibitem[\protect\citeauthoryear{Izadi}{Izadi}{2020}]%
        {Izadi2020generalized}
\bibfield{author}{\bibinfo{person}{Farzali Izadi}.}
  \bibinfo{year}{2020}\natexlab{}.
\newblock \showarticletitle{Generalized additive models to capture the death
  rates in Canada COVID-19}.
\newblock \bibinfo{journal}{\emph{arXiv preprint arXiv:1702.08608}}
  (\bibinfo{year}{2020}).
\newblock


\bibitem[\protect\citeauthoryear{Johnson, Pollard, Shen, Li-Wei, Feng,
  Ghassemi, Moody, Szolovits, Celi, and Mark}{Johnson et~al\mbox{.}}{2016}]%
        {johnson2016mimic}
\bibfield{author}{\bibinfo{person}{Alistair~EW Johnson}, \bibinfo{person}{Tom~J
  Pollard}, \bibinfo{person}{Lu Shen}, \bibinfo{person}{H~Lehman Li-Wei},
  \bibinfo{person}{Mengling Feng}, \bibinfo{person}{Mohammad Ghassemi},
  \bibinfo{person}{Benjamin Moody}, \bibinfo{person}{Peter Szolovits},
  \bibinfo{person}{Leo~Anthony Celi}, {and} \bibinfo{person}{Roger~G Mark}.}
  \bibinfo{year}{2016}\natexlab{}.
\newblock \showarticletitle{MIMIC-III, a freely accessible critical care
  database}.
\newblock \bibinfo{journal}{\emph{Scientific Data}} \bibinfo{volume}{3},
  \bibinfo{number}{1} (\bibinfo{year}{2016}).
\newblock


\bibitem[\protect\citeauthoryear{Lipton}{Lipton}{2018}]%
        {lipton2018mythos}
\bibfield{author}{\bibinfo{person}{Zachary~C Lipton}.}
  \bibinfo{year}{2018}\natexlab{}.
\newblock \showarticletitle{The mythos of model interpretability}.
\newblock \bibinfo{journal}{\emph{Queue}} \bibinfo{volume}{16},
  \bibinfo{number}{3} (\bibinfo{year}{2018}).
\newblock


\bibitem[\protect\citeauthoryear{Lou, Caruana, and Gehrke}{Lou
  et~al\mbox{.}}{2012}]%
        {lou2012intelligible}
\bibfield{author}{\bibinfo{person}{Yin Lou}, \bibinfo{person}{Rich Caruana},
  {and} \bibinfo{person}{Johannes Gehrke}.} \bibinfo{year}{2012}\natexlab{}.
\newblock \showarticletitle{Intelligible models for classification and
  regression}. In \bibinfo{booktitle}{\emph{KDD}}.
\newblock


\bibitem[\protect\citeauthoryear{Lundberg and Lee}{Lundberg and Lee}{2017}]%
        {lundberg2017unified}
\bibfield{author}{\bibinfo{person}{Scott~M Lundberg} {and}
  \bibinfo{person}{Su-In Lee}.} \bibinfo{year}{2017}\natexlab{}.
\newblock \showarticletitle{A Unified Approach to Interpreting Model
  Predictions}. In \bibinfo{booktitle}{\emph{NeurIPS}}.
\newblock


\bibitem[\protect\citeauthoryear{Mehrabi, Morstatter, Saxena, Lerman, and
  Galstyan}{Mehrabi et~al\mbox{.}}{2019}]%
        {mehrabi2019survey}
\bibfield{author}{\bibinfo{person}{Ninareh Mehrabi}, \bibinfo{person}{Fred
  Morstatter}, \bibinfo{person}{Nripsuta Saxena}, \bibinfo{person}{Kristina
  Lerman}, {and} \bibinfo{person}{Aram Galstyan}.}
  \bibinfo{year}{2019}\natexlab{}.
\newblock \showarticletitle{A survey on bias and fairness in machine learning}.
\newblock \bibinfo{journal}{\emph{arXiv preprint arXiv:1908.09635}}
  (\bibinfo{year}{2019}).
\newblock


\bibitem[\protect\citeauthoryear{Munson and Caruana}{Munson and
  Caruana}{2009}]%
        {munson2009feature}
\bibfield{author}{\bibinfo{person}{M~Arthur Munson} {and} \bibinfo{person}{Rich
  Caruana}.} \bibinfo{year}{2009}\natexlab{}.
\newblock \showarticletitle{On feature selection, bias-variance, and bagging}.
  In \bibinfo{booktitle}{\emph{ECML PKDD}}.
\newblock


\bibitem[\protect\citeauthoryear{Nori, Jenkins, Koch, and Caruana}{Nori
  et~al\mbox{.}}{2019}]%
        {nori2019interpretml}
\bibfield{author}{\bibinfo{person}{Harsha Nori}, \bibinfo{person}{Samuel
  Jenkins}, \bibinfo{person}{Paul Koch}, {and} \bibinfo{person}{Rich Caruana}.}
  \bibinfo{year}{2019}\natexlab{}.
\newblock \showarticletitle{InterpretML: A Unified Framework for Machine
  Learning Interpretability}.
\newblock \bibinfo{journal}{\emph{arXiv preprint arXiv:1909.09223}}
  (\bibinfo{year}{2019}).
\newblock


\bibitem[\protect\citeauthoryear{Pedersen, Miller, Simpson, and Ross}{Pedersen
  et~al\mbox{.}}{2019}]%
        {pedersen2019hierarchical}
\bibfield{author}{\bibinfo{person}{Eric~J Pedersen}, \bibinfo{person}{David~L
  Miller}, \bibinfo{person}{Gavin~L Simpson}, {and} \bibinfo{person}{Noam
  Ross}.} \bibinfo{year}{2019}\natexlab{}.
\newblock \showarticletitle{Hierarchical generalized additive models in
  ecology: an introduction with mgcv}.
\newblock \bibinfo{journal}{\emph{PeerJ}}  \bibinfo{volume}{7}
  (\bibinfo{year}{2019}).
\newblock


\bibitem[\protect\citeauthoryear{Petersen, Witten, and Simon}{Petersen
  et~al\mbox{.}}{2016}]%
        {petersen2016fused}
\bibfield{author}{\bibinfo{person}{Ashley Petersen}, \bibinfo{person}{Daniela
  Witten}, {and} \bibinfo{person}{Noah Simon}.}
  \bibinfo{year}{2016}\natexlab{}.
\newblock \showarticletitle{Fused lasso additive model}.
\newblock \bibinfo{journal}{\emph{Journal of Computational and Graphical
  Statistics}} \bibinfo{volume}{25}, \bibinfo{number}{4}
  (\bibinfo{year}{2016}).
\newblock


\bibitem[\protect\citeauthoryear{Ravindra, Rattan, Mor, and Aggarwal}{Ravindra
  et~al\mbox{.}}{2019}]%
        {ravindra2019generalized}
\bibfield{author}{\bibinfo{person}{Khaiwal Ravindra}, \bibinfo{person}{Preety
  Rattan}, \bibinfo{person}{Suman Mor}, {and} \bibinfo{person}{Ashutosh~Nath
  Aggarwal}.} \bibinfo{year}{2019}\natexlab{}.
\newblock \showarticletitle{Generalized additive models: Building evidence of
  air pollution, climate change and human health}.
\newblock \bibinfo{journal}{\emph{Environment International}}
  \bibinfo{volume}{132} (\bibinfo{year}{2019}).
\newblock


\bibitem[\protect\citeauthoryear{Ribeiro, Singh, and Guestrin}{Ribeiro
  et~al\mbox{.}}{2016}]%
        {lime}
\bibfield{author}{\bibinfo{person}{Marco~Tulio Ribeiro},
  \bibinfo{person}{Sameer Singh}, {and} \bibinfo{person}{Carlos Guestrin}.}
  \bibinfo{year}{2016}\natexlab{}.
\newblock \showarticletitle{``Why Should I Trust You?": Explaining the
  Predictions of Any Classifier}. In \bibinfo{booktitle}{\emph{KDD}}.
\newblock


\bibitem[\protect\citeauthoryear{Rostami, Simbar, Amiri, Bidhendi-Yarandi,
  Hosseinpanah, and Tehrani}{Rostami et~al\mbox{.}}{2020}]%
        {rostami2020optimal}
\bibfield{author}{\bibinfo{person}{Maryam Rostami}, \bibinfo{person}{Masoumeh
  Simbar}, \bibinfo{person}{Mina Amiri}, \bibinfo{person}{Razieh
  Bidhendi-Yarandi}, \bibinfo{person}{Farhad Hosseinpanah}, {and}
  \bibinfo{person}{Fahimeh~Ramezani Tehrani}.} \bibinfo{year}{2020}\natexlab{}.
\newblock \showarticletitle{The optimal cut-off point of vitamin D for
  pregnancy outcomes using a generalized additive model}.
\newblock \bibinfo{journal}{\emph{Clinical Nutrition}} (\bibinfo{year}{2020}).
\newblock


\bibitem[\protect\citeauthoryear{Sadhanala and Tibshirani}{Sadhanala and
  Tibshirani}{2019}]%
        {sadhanala2017additive}
\bibfield{author}{\bibinfo{person}{Veeranjaneyulu Sadhanala} {and}
  \bibinfo{person}{Ryan~J Tibshirani}.} \bibinfo{year}{2019}\natexlab{}.
\newblock \showarticletitle{Additive models with trend filtering}.
\newblock \bibinfo{journal}{\emph{The Annals of Statistics}}
  (\bibinfo{year}{2019}).
\newblock


\bibitem[\protect\citeauthoryear{Sapra}{Sapra}{2013}]%
        {sapra2013generalized}
\bibfield{author}{\bibinfo{person}{K Sapra}.} \bibinfo{year}{2013}\natexlab{}.
\newblock \showarticletitle{Generalized additive models in business and
  economics}.
\newblock \bibinfo{journal}{\emph{International Journal of Advanced Statistics
  and Probability}} \bibinfo{volume}{1}, \bibinfo{number}{3}
  (\bibinfo{year}{2013}).
\newblock


\bibitem[\protect\citeauthoryear{Saw, Moser, Martens, and Franceschi}{Saw
  et~al\mbox{.}}{2017}]%
        {saw2017applying}
\bibfield{author}{\bibinfo{person}{Nay Min Min~Thaw Saw},
  \bibinfo{person}{Claudio Moser}, \bibinfo{person}{Stefan Martens}, {and}
  \bibinfo{person}{Pietro Franceschi}.} \bibinfo{year}{2017}\natexlab{}.
\newblock \showarticletitle{Applying generalized additive models to unravel
  dynamic changes in anthocyanin biosynthesis in methyl jasmonate elicited
  grapevine (Vitis vinifera cv. Gamay) cell cultures}.
\newblock \bibinfo{journal}{\emph{Horticulture Research}} \bibinfo{volume}{4},
  \bibinfo{number}{1} (\bibinfo{year}{2017}).
\newblock


\bibitem[\protect\citeauthoryear{Servén and Brummitt}{Servén and
  Brummitt}{2018}]%
        {pygam}
\bibfield{author}{\bibinfo{person}{Daniel Servén} {and}
  \bibinfo{person}{Charlie Brummitt}.} \bibinfo{year}{2018}\natexlab{}.
\newblock \bibinfo{title}{pyGAM: Generalized Additive Models in Python}.
\newblock
\newblock
\urldef\tempurl%
\url{https://doi.org/10.5281/zenodo.1208723}
\showDOI{\tempurl}


\bibitem[\protect\citeauthoryear{Shmueli}{Shmueli}{2010}]%
        {shmueli2010explain}
\bibfield{author}{\bibinfo{person}{Galit Shmueli}.}
  \bibinfo{year}{2010}\natexlab{}.
\newblock \showarticletitle{To explain or to predict?}
\newblock \bibinfo{journal}{\emph{Statist. Sci.}} \bibinfo{volume}{25},
  \bibinfo{number}{3} (\bibinfo{year}{2010}).
\newblock


\bibitem[\protect\citeauthoryear{Slack, Hilgard, Jia, Singh, and
  Lakkaraju}{Slack et~al\mbox{.}}{2020}]%
        {slack2020fooling}
\bibfield{author}{\bibinfo{person}{Dylan Slack}, \bibinfo{person}{Sophie
  Hilgard}, \bibinfo{person}{Emily Jia}, \bibinfo{person}{Sameer Singh}, {and}
  \bibinfo{person}{Himabindu Lakkaraju}.} \bibinfo{year}{2020}\natexlab{}.
\newblock \showarticletitle{Fooling LIME and SHAP: Adversarial Attacks on Post
  hoc Explanation Methods}. In \bibinfo{booktitle}{\emph{AIES}}.
\newblock


\bibitem[\protect\citeauthoryear{Strobl, Boulesteix, Zeileis, and
  Hothorn}{Strobl et~al\mbox{.}}{2007}]%
        {strobl2007bias}
\bibfield{author}{\bibinfo{person}{Carolin Strobl}, \bibinfo{person}{Anne-Laure
  Boulesteix}, \bibinfo{person}{Achim Zeileis}, {and} \bibinfo{person}{Torsten
  Hothorn}.} \bibinfo{year}{2007}\natexlab{}.
\newblock \showarticletitle{Bias in random forest variable importance measures:
  Illustrations, sources and a solution}.
\newblock \bibinfo{journal}{\emph{BMC Bioinformatics}} \bibinfo{volume}{8},
  \bibinfo{number}{1} (\bibinfo{year}{2007}).
\newblock


\bibitem[\protect\citeauthoryear{Tan, Caruana, Hooker, Koch, and Gordo}{Tan
  et~al\mbox{.}}{2018b}]%
        {tan2018learning}
\bibfield{author}{\bibinfo{person}{Sarah Tan}, \bibinfo{person}{Rich Caruana},
  \bibinfo{person}{Giles Hooker}, \bibinfo{person}{Paul Koch}, {and}
  \bibinfo{person}{Albert Gordo}.} \bibinfo{year}{2018}\natexlab{b}.
\newblock \showarticletitle{Learning global additive explanations for neural
  nets using model distillation}.
\newblock \bibinfo{journal}{\emph{arXiv preprint arXiv:1801.08640}}
  (\bibinfo{year}{2018}).
\newblock


\bibitem[\protect\citeauthoryear{Tan, Caruana, Hooker, and Lou}{Tan
  et~al\mbox{.}}{2018a}]%
        {tan2018distill}
\bibfield{author}{\bibinfo{person}{Sarah Tan}, \bibinfo{person}{Rich Caruana},
  \bibinfo{person}{Giles Hooker}, {and} \bibinfo{person}{Yin Lou}.}
  \bibinfo{year}{2018}\natexlab{a}.
\newblock \showarticletitle{Distill-and-compare: Auditing black-box models
  using transparent model distillation}. In \bibinfo{booktitle}{\emph{AIES}}.
\newblock


\bibitem[\protect\citeauthoryear{Tibshirani}{Tibshirani}{1996}]%
        {tibshirani1996regression}
\bibfield{author}{\bibinfo{person}{Robert Tibshirani}.}
  \bibinfo{year}{1996}\natexlab{}.
\newblock \showarticletitle{Regression shrinkage and selection via the lasso}.
\newblock \bibinfo{journal}{\emph{Journal of the Royal Statistical Society:
  Series B}} \bibinfo{volume}{58}, \bibinfo{number}{1} (\bibinfo{year}{1996}).
\newblock


\bibitem[\protect\citeauthoryear{Tibshirani, Saunders, Rosset, Zhu, and
  Knight}{Tibshirani et~al\mbox{.}}{2005}]%
        {fusedlasso}
\bibfield{author}{\bibinfo{person}{Robert Tibshirani}, \bibinfo{person}{Michael
  Saunders}, \bibinfo{person}{Saharon Rosset}, \bibinfo{person}{Ji Zhu}, {and}
  \bibinfo{person}{Keith Knight}.} \bibinfo{year}{2005}\natexlab{}.
\newblock \showarticletitle{Sparsity and smoothness via the fused lasso}.
\newblock \bibinfo{journal}{\emph{Journal of the Royal Statistical Society:
  Series B}} \bibinfo{volume}{67}, \bibinfo{number}{1} (\bibinfo{year}{2005}).
\newblock


\bibitem[\protect\citeauthoryear{Wahba}{Wahba}{1985}]%
        {wahba1985comparison}
\bibfield{author}{\bibinfo{person}{Grace Wahba}.}
  \bibinfo{year}{1985}\natexlab{}.
\newblock \showarticletitle{A comparison of GCV and GML for choosing the
  smoothing parameter in the generalized spline smoothing problem}.
\newblock \bibinfo{journal}{\emph{The Annals of Statistics}}
  (\bibinfo{year}{1985}).
\newblock


\bibitem[\protect\citeauthoryear{Wahba}{Wahba}{1990}]%
        {wahba1990spline}
\bibfield{author}{\bibinfo{person}{Grace Wahba}.}
  \bibinfo{year}{1990}\natexlab{}.
\newblock \bibinfo{booktitle}{\emph{Spline models for observational data}}.
\newblock \bibinfo{publisher}{SIAM}.
\newblock


\bibitem[\protect\citeauthoryear{Wood}{Wood}{2011}]%
        {mgcv}
\bibfield{author}{\bibinfo{person}{S.~N. Wood}.}
  \bibinfo{year}{2011}\natexlab{}.
\newblock \showarticletitle{Fast stable restricted maximum likelihood and
  marginal likelihood estimation of semiparametric generalized linear models}.
\newblock \bibinfo{journal}{\emph{Journal of the Royal Statistical Society:
  Series B}} \bibinfo{volume}{73}, \bibinfo{number}{1} (\bibinfo{year}{2011}).
\newblock


\bibitem[\protect\citeauthoryear{Wright and K{\"o}nig}{Wright and
  K{\"o}nig}{2019}]%
        {wright2019splitting}
\bibfield{author}{\bibinfo{person}{Marvin~N Wright} {and}
  \bibinfo{person}{Inke~R K{\"o}nig}.} \bibinfo{year}{2019}\natexlab{}.
\newblock \showarticletitle{Splitting on categorical predictors in random
  forests}.
\newblock \bibinfo{journal}{\emph{PeerJ}}  \bibinfo{volume}{7}
  (\bibinfo{year}{2019}).
\newblock


\bibitem[\protect\citeauthoryear{Zemel, Wu, Swersky, Pitassi, and Dwork}{Zemel
  et~al\mbox{.}}{2013}]%
        {zemel2013learning}
\bibfield{author}{\bibinfo{person}{Rich Zemel}, \bibinfo{person}{Yu Wu},
  \bibinfo{person}{Kevin Swersky}, \bibinfo{person}{Toni Pitassi}, {and}
  \bibinfo{person}{Cynthia Dwork}.} \bibinfo{year}{2013}\natexlab{}.
\newblock \showarticletitle{Learning fair representations}. In
  \bibinfo{booktitle}{\emph{ICML}}.
\newblock


\bibitem[\protect\citeauthoryear{Zhou and Hooker}{Zhou and Hooker}{2021}]%
        {zhou2019unbiased}
\bibfield{author}{\bibinfo{person}{Zhengze Zhou} {and} \bibinfo{person}{Giles
  Hooker}.} \bibinfo{year}{2021}\natexlab{}.
\newblock \showarticletitle{Unbiased Measurement of Feature Importance in
  Tree-Based Methods}.
\newblock \bibinfo{journal}{\emph{TKDD}} (\bibinfo{year}{2021}).
\newblock


\end{thebibliography}
\end{small}

% \newpage

\appendix

\section{Reproducibility: Training details, hyperparameters, and datasets}
\label{sec:appx_training_details}

Code can be found at \url{https://github.com/zzzace2000/GAMs}.

\subsection{Further training details and hyperparameters}
In this section, we further describe training details and hyperparameters to supplement the discussion in Sec. \ref{sec:gam_intro}.

\begin{itemize}[leftmargin=*]
    \item EBM, EBM-BF: we use the open-source package from \url{https://github.com/interpretml/interpret}. We set the parameters inner bagging as $100$ and outer bagging as $100$. We find that increasing the number of bags does not further improve performance. We use the default learning rate of $0.01$, default early stopping patience set to $50$, and the maximum $30000$ episodes to make sure it converges.
    \item XGB, XGB-d3, XGB-L2: we use the open source package  \url{https://xgboost.readthedocs.io/en/latest/index.html}. We also use the default learning rate with the same early stopping patience set as $50$ and number of trees as maximum $30,000$. We use bagging of $100$ times and depth $1$ for our XGB GAM model.
    For XGB-d3 (XGB with tree depth $3$), we find that bagging of XGB-d3 hurts the performance a bit, and thus do not apply any bagging for XGB-d3. For XGB-L2, we set the parameter "colsample\_bytree" as a small value 1e-5 to make sure each tree only sees one feature.
    \item FLAM: we use the package from R \url{https://cran.r-project.org/web/packages/flam/flam.pdf}. We use a $15\%$ validation set to select the best $\lambda$ penalty parameter in the fused LASSO, and then refit the whole data with the best penalty parameter.
    We set the parameter number of lambda as $100$ and the minimum ratio as 1e-4 to increase the performance of the model. 
    \item Spline: we use the pygam package~\citep{pygam}. We set the number of basis functions to be $50$ and the maximum iteration as $500$. We find increasing number of basis functions more than $50$ would result in instability when fitting in large datasets.
    \item LR: we use scikit-learn's LogisticRegressionCV with $Cs=12$ (grid search for $12$ different $\ell_2$ penalty) and cross validation for $5$ times to choose the best $\ell_2$, and re-fit on the whole data.
    \item iLR, mLR: we use the EBM package's preprocessor to quantily bin the features into $255$ bins. Then we use LR on top of it to train a linear model.
\end{itemize}

We also tried the following GAM algorithms but do not include them in the main results, for reasons detailed below:
\begin{itemize}[leftmargin=*]
    \item SKGBT: we try the gradient boosting tree in scikit-learn also with tree depth set as $1$. The result is similar to EBM so we do not compare them in the paper.
    \item Cubic spline and plate spline in R mgcv package: to our surprise, mgcv is really unstable on two datasets, Breast Cancer and Churn. After some investigation, we found a possible reason to be that mgcv does not handle numerical instability when the prediction is too close to $0$ or $1$.
    
\end{itemize}

\subsection{Encoding categorical features}

For datasets with categorical variables, the choice of encoding can affect both the shape plots and the accuracy. 
For gradient boosting trees, one may think that using label encoding (LE) is better than one-hot encoding, as one-hot encoding has been shown to have inferior performance in ensemble trees \citep{wright2019splitting}.
% In Table \ref{table:encoding}, 
We investigate the effects of two types of encoding on EBM and XGB. 
In $6$ of the datasets with categorical features, EBM with label encoding (LE) indeed shows superior performance to one-hot encoding.
However, for XGB, one-hot encoding performs slightly better on average.
% We also qualitatively examine one of the categorical features (Race) in the COMPAS dataset. 
% In Fig.~\ref{fig:encoding_compass_race}, 
% we find that EBM shape plots have similar patterns whether LE or one-hot encoding is used, but XGB with LE gives much smaller effect than one-hot encoding in the category Asian, suggesting that the order of the features affects its estimate.
Thus we use LE for EBM and one-hot encoding for XGB.
For the rest of the methods, we use LE for mLR and one-hot encoding for FLAM, Spline, LR and iLR as these methods cannot handle inadequate numerical ordering.

\subsection{Dataset sources}
The datasets used in this paper can be found at: 
\begin{itemize}[leftmargin=*]
    \item Adult: UCI \cite{UCI}
    \item Breast cancer: UCI \cite{UCI}
    \item Credit: \url{https://www.kaggle.com/mlg-ulb/creditcardfraud}
    \item Churn: \url{https://www.kaggle.com/blastchar/telco-customer-churn}
    \item COMPAS: \url{https://www.kaggle.com/danofer/compass}
    \item Heart disease: UCI \cite{UCI}
    \item MIMIC-II and MIMIC-III dataset \citep{johnson2016mimic}
    \item Pneumonia: we thank the authors of \citet{caruana2015intelligible} for running our code on their dataset. 
    \item Support2: \url{http://biostat.mc.vanderbilt.edu/DataSets}
\end{itemize}

\section{Additional shape plots}
\label{sec:appx_additional_shape_graphs}
The complete set of shape plots can be found at \newline 
\url{https://drive.google.com/file/d/1PoMRgfuHYax6xuCVU0Dbut3yFJ2ohuLX/view?usp=sharing}.

\end{document}